\documentclass[10pt,conference]{IEEEtran}
\IEEEoverridecommandlockouts
\usepackage{cite}
\usepackage{amsmath,amssymb,amsfonts}
\usepackage{graphicx}
\usepackage{textcomp}
\usepackage{xcolor}

 \usepackage{colortbl}
\usepackage{mathrsfs} 
\usepackage{color}
\usepackage{bm}
\usepackage{algorithm}  
\usepackage[noend]{algpseudocode}
\usepackage{url}
\usepackage{booktabs} 
\usepackage{makecell}
\usepackage{threeparttable}
\usepackage{multirow}
\usepackage{framed}
\usepackage{array}
\usepackage{subfig}
\newtheorem{problem}{Problem} 
\newtheorem{definition}{Definition}

\newcommand \bird {BIRDNN}

\def\BibTeX{{\rm B\kern-.05em{\sc i\kern-.025em b}\kern-.08em
    T\kern-.1667em\lower.7ex\hbox{E}\kern-.125emX}}
\begin{document}

\title{Repairing Deep Neural Networks Based on Behavior Imitation}

\author{\IEEEauthorblockN{Zhen Liang\IEEEauthorrefmark{1}\IEEEauthorrefmark{5}, Taoran Wu\IEEEauthorrefmark{2}\IEEEauthorrefmark{3}\IEEEauthorrefmark{5}, Changyuan Zhao\IEEEauthorrefmark{2}\IEEEauthorrefmark{3}, Wanwei Liu\IEEEauthorrefmark{4}$^{\#}$,  Bai Xue\IEEEauthorrefmark{2}, Wenjing Yang \IEEEauthorrefmark{1} and Ji Wang\IEEEauthorrefmark{1}} \IEEEauthorblockA{\IEEEauthorrefmark{1}Institute for Quantum Information \& State Key Laboratory of High Performance Computing,\\ National University of Defense Technology, Changsha, China \\
Email: \{liangzhen, wenjing.yang, wj\}@nudt.edu.cn} \IEEEauthorblockA{\IEEEauthorrefmark{2}State Key Laboratory of Computer Science Institute of Software,  Chinese Academy of Sciences, Beijing, China\\
Email: \{wutr, zhaocy, xuebai\}@ios.ac.cn}\IEEEauthorblockA{\IEEEauthorrefmark{3}University of Chinese Academy of Sciences, Beijing, China} \IEEEauthorblockA{\IEEEauthorrefmark{4}College of Computer Science and Technology, National University of Defense
Technology, Changsha, China \\
Email: wwliu@nudt.edu.cn}
}

\maketitle

\begin{abstract}
The increasing use of deep neural networks (DNNs) in safety-critical systems has raised concerns about their potential for exhibiting ill-behaviors. While DNN verification and testing provide post hoc conclusions regarding unexpected behaviors, they do not prevent the erroneous behaviors from occurring. To address this issue, DNN repair/patch aims to eliminate unexpected predictions generated by defective DNNs. Two typical DNN repair paradigms are retraining and fine-tuning. However, existing methods focus on the high-level abstract interpretation or inference of state spaces, ignoring the underlying neurons' outputs. This renders patch processes computationally prohibitive and limited to piecewise linear (PWL) activation functions to great extent. To address these shortcomings, we propose a behavior-imitation based repair framework, BIRDNN, which integrates the two repair paradigms for the first time. BIRDNN corrects incorrect predictions of negative samples by imitating the closest expected behaviors of positive samples during the retraining repair procedure. For the fine-tuning repair process, BIRDNN analyzes the behavior differences of neurons on positive and negative samples to identify the most responsible neurons for the erroneous behaviors. To tackle more challenging domain-wise repair problems (DRPs), we synthesize BIRDNN with a domain behavior characterization technique to repair buggy DNNs in a probably approximated correct style. We also implement a prototype tool based on BIRDNN and evaluate it on ACAS Xu DNNs. Our experimental results show that BIRDNN can successfully repair buggy DNNs with significantly higher efficiency than state-of-the-art repair tools. Additionally, BIRDNN is highly compatible with different activation functions.

\end{abstract}

\begin{IEEEkeywords}
DNN repair/patch, behavior imitation, retraining, fine-tuning, fault localization
\end{IEEEkeywords}


\section{Introduction}

Deep neural networks (DNNs) have become a leading candidate computation model for deep learning in recent years and have achieved enormous success in various domains, including computer vision \cite{dahnert2021panoptic} and natural language processing \cite{yuan2021bartscore,karch2021grounding}. Despite their popularity and success, DNNs are not infallible and can produce erroneous outputs, such as incorrect predictions, classifications, or identifications. These unexpected errors can lead to system failures\cite{DBLP:journals/asc/XuXTH03, DBLP:conf/IEEEicci/Berwick21,DBLP:conf/latw/BosioBR019}, wrongful arrests \cite{translationarrest, wronglyaccused}, and even loss of life \cite{teslacrash}. As a result, the unexpected behaviors of DNNs make their application in safety-critical domains, such as autonomous vehicles, a significant and pressing concern.

To earn users' trust and alleviate their concerns, it is critical to ensure that DNNs meet specific requirements and produce expected outputs before deployment. Recent advancements in \emph{DNN verification}, \emph{DNN testing}, and \emph{DNN property-guided training}  have highlighted the importance of this issue. Diverse DNN verification techniques \cite{2021Enhancing,DBLP:journals/jcst/LiuSZW20} have been proposed to verify whether a given DNN adheres to specific properties such as robustness\cite{goodfellow2014explaining} or safety\cite{liang2022safety}. These techniques range from abstract interpretation \cite{DBLP:journals/pacmpl/SinghGPV19,DBLP:conf/tacas/YangLLHWSXZ21,DBLP:conf/sp/GehrMDTCV18,DBLP:conf/icml/KoLWDWL19}, SMT/SAT theory \cite{DBLP:conf/cav/KatzBDJK17,DBLP:conf/tacas/Amir0BK21} to linear programming \cite{DBLP:conf/iclr/TjengXT19,DBLP:conf/ijcai/BattenKLZ21,DBLP:conf/cvpr/LinYCZLLH19}. Meanwhile, DNN testing \cite{DBLP:conf/qrs/ShenWC18,DBLP:conf/icse/TianPJR18} evaluates the behaviors of DNNs through numerous test cases and has made significant progress in coverage criteria \cite{DBLP:conf/issta/XieMJXCLZLYS19} and case generation \cite{DBLP:conf/icse/ZhangXMDH0Z020,DBLP:journals/corr/abs-2005-00760} recently. However, both verification and testing are post hoc analyses, and their qualitative or quantitative results do not provide comfort when properties are violated. Conversely, property-guided training methods \cite{gowal2018effectiveness,madry2017towards} focus on training DNNs that are correct-by-construction during the training phase. Moreover, much attention is being paid to training methods for developing robust DNNs.

Although pre-trained DNNs are highly effective in many applications, they can struggle to handle erroneous behaviors. This leads us to the main topic of our paper: DNN repair, which is also sometimes called DNN patching. When a DNN is found to be behaving poorly (as detected by DNN verification or testing), DNN repair or patching aims to eliminate these unexpected behaviors while minimizing the impact on performance. There are currently two main repair paradigms.


On the one hand, DNN retraining is a powerful tool for correcting erroneous outputs of DNNs and is similar to property-guided training. Retraining-based methods \cite{yang2021reachability, sinitsin2020editable} remove mistakes by evolving existing buggy DNNs, while property-guided training avoids specification violations from the beginning. The key to successful DNN retraining is balancing the need to maintain original performance while eliminating unexpected behaviors, achieved by using appropriate loss functions. However, retraining can be computationally expensive, and original training datasets are not always publicly available. Additionally, undifferentiated parameter updates during retraining can introduce new bad behaviors.

On the other hand, unlike the global and aimless parameter modification in DNN retraining, \emph{DNN fine-tuning} \cite{tao2023architecture} devotes to repairing buggy DNNs locally, intentionally and slightly, i.e., patching a specific subset of the DNN parameters to get rid of erroneous predictions. Initially, fine-tuning based methods\cite{GoldbergerKAK20,sotoudeh2021provable} typically convert the adjustment of DNN parameters located in a 
DNN layer (arbitrary or well-selected) into a constraint solving problem of parameters and  minimize parameter modifications then. Compared to the adjustment of parameters on a single layer, some recent work has taken fault localization into account\cite{DBLP:conf/cav/UsmanGSNP21,sun2022causality}, which is utilized to determine a parameter subset that has a more significant impact on error behaviors, to make fine-tuning processes more intentionally. The fine-tuning methods generally outperform the retraining ones due to their slight and local modifications.

When considering either the retraining based or the fine-tuning based DNN repair methods, current attention is primarily directed towards high-level abstract interpretation or inference of state spaces of DNN layers, while neglecting behaviors of the underlying neurons. This leads to existing repair methods being computationally prohibitive and restricting the range of repaired DNNs. For instance, repairing DNNs through linear mapping regions necessitates the expensive calculation of the sets of polytope abstract domains, and non-piecewise linear activation functions are limited in the effectiveness.

To overcome these dilemmas, this paper proposes a behavior-imitation based DNN repair framework, \bird \  (\textbf{B}ehavior-\textbf{I}mitation Based \textbf{R}epair of \textbf{DNN}s).  In essence, we investigate the (hidden) states of neurons within DNNs, analyzing neurons' behavior differences between the expected behaviors on positive samples and the unexpected behaviors on negative samples. Resorting to behavior imitation, we present alternative retraining based and fine-tuning based repair methods to patch defective DNNs, and \bird \ is the first repair framework unifying DNN retraining and DNN fine-tuning together. For the retraining based method, we assign new correct labels to negative sample inputs, imitating the outputs of nearby positive samples, and then retrain original DNNs to improve erroneous behavior. For the fine-tuning based method, fault localization identifies the most responsible neurons for unexpected predictions by analyzing behavior differences on positive and negative samples. Then we utilize particle swarm optimization (PSO) to modify those ``guilty" neurons to maintain the original DNN performance while minimizing property violations. Our paper aims to tackle the more difficult domain-wise repair problems (DRPs) and to this end,  we also integrate \bird \ with a sampling based technique to character DNN behaviors over domains, repairing buggy DNNs in a probably approximated correct style.

\textbf{Contributions.} Main contributions of this paper are listed as follows.
\begin{itemize}
    \item Based on the investigation on neuron behaviors and the insight of behavior imitation, we propose a novel DNN repair framework, \bird, which supports alternative retraining style and fine-tuning style repair simultaneously for the first time, to our best knowledge.
    
    \item Within \bird, DNNs imitate the expected behaviors of the closest positive samples when encountering negative inputs during retraining procedures. For the fine-tuning method, \bird \ adopts a straightforward and efficient fault localization strategy on behavior difference analysis to identify the most responsible neurons.
    
    \item To tackle the more challenging DRPs, we integrate \bird \ with a characterization technique of DNN domain behaviors, reducing DRPs to sample-based repair problems and repairing buggy DNNs in a probably approximated correct style.
    
    \item We have implemented a prototype tool based on \bird, which is available from {\url{https://github.com/ByteTao5/BIRDNN}}. The experimental results on ACAS Xu DNNs illustrate the effectiveness, efficiency and compatibility of our proposed DNN patch framework. 
\end{itemize}

The remainder of this paper is organized as follows. Section \ref{prelimi} provides an introduction of necessary background. Afterward, Section \ref{problem} formulates the concerned and addressed domain-wise repair problems. Following this, Section \ref{method} detailedly demonstrates our behavior-imitation based DNN repair framework, \bird. Next, Section \ref{exp} implements a prototype tool and reports its effectiveness, efficiency and compatibility on the widely-used ACAS Xu DNNs versus state-of-the-art methods.  Section \ref{related} supplements some related work and comparisons. Finally, Section \ref{concl} summarizes this paper and gives possible future directions.

\section{Preliminaries}
\label{prelimi}

When training or testing DNNs, complex data flows arise and property specifications impose specific requirements on the generated data flows. In the following, we present some preliminaries related to DNNs and property specifications.

\subsection{Deep Neural Networks}
An  $L$-layer DNN usually contains an input layer, an output layer, and $L-2$ successive hidden layers. A DNN specified with the parameter set $\theta$ is denoted by $\mathcal{N}_{\theta}:\mathbb{R}^{m}\rightarrow \mathbb{R}^{n}$, containing $m$ and $n$ neurons in its input and output layers, which are termed the input/output dimensions respectively. That is to say, the layer dimension $d_i$ refers to the number of neurons located on the $i$-th DNN layer. Beginning with an input (or, sample) $\bm{x}\in \mathbb{R}^{m}$ on the input layer, it comes into the \textit{forward propagation}\cite{mitchell2007machine}. Between two adjacent DNN layers, a non-linear activation function follows an affine transformation, whereas only an affine transformation exists for the last two layers. 
The common activation functions are \texttt{ReLU}, \texttt{tanh} and \texttt{Sigmoid}, where the first one is piecewise linear (PWL) and the others are strictly non-linear. Generally, PWL functions are easier to tackle due the nice algebraic property.  The whole forward propagation process is the composition of the operations between each two adjacent layers and it generates the output (or, prediction) $\bm{y}:=\mathcal{N}_\theta(\bm{x})\in \mathbb{R}^{n}$ with respect to the sample $\bm{x}$.

There appears a data flow in DNN $\mathcal{N}_\theta$ when 
processing the input $\bm{x}$. We name the data flow as the \textit{behavior} of the DNN  $\mathcal{N}_\theta$ on the sample $\bm{x}$ and it consists of the input vector, hidden states for each hidden layer and the output vector. The DNN behaviour on sample $\bm{x}$ can be represented with $\{\mathcal{N}^{0}_{\theta}(\bm{x})$ (i.e., $\bm{x}$), $ \mathcal{N}^{1}_{\theta}(\bm{x})$, $\cdots,$ $\mathcal{N}^{L-1}_{\theta}$ (i.e., $\bm{y}$ or $\mathcal{N}_\theta(\bm{x})$)$\}$. 

\subsection{Property Specifications}
Before the deployment in practical applications, DNNs are required to satisfy certain properties, such as robustness\cite{goodfellow2014explaining,casadio2022neural}, safety\cite{liang2022safety}, reachability\cite{huang2019reachnn,yang2021reachability}, fairness\cite{sun2021probabilistic} and so on, especially in safety-critical domains.  Consequently, property specifications emerge as the times demand and they are widely utilized in DNN property-guided training\cite{fischer2019dl2} and DNN verification\cite{liu2020verifying} related fields.

A property specification $\mathcal{P}$ w.r.t. a DNN $\mathcal{N}_{\theta}$ generally consists of two components, a \textit{pre-condition} $\varphi$ and a \textit{post-condition} $\psi$, i.e., $\mathcal{P}:=\{\varphi;\psi\}$. The pre-condition $\varphi$ is a constraint on the input domain $\mathbb{R}^{m}$ and it describes a \textit{predetermined input set} of $\mathcal{N}_{\theta}$. Likewise, the post-condition $\psi$ restricts the output domain $\mathbb{R}^{n}$, representing a \textit{desired output region}. Then, the specification $\mathcal{P}$ imposes a behavior requirement for $\mathcal{N}_{\theta}$ that it should ideally map the predetermined input set into the desired output region (i.e, expected behaviors), or the DNN is buggy. Further, the relation $\varphi \stackrel{\mathcal{N}_{\theta}}\rightarrow \psi$ refers that DNN $\mathcal{N}_{\theta}$ satisfies the property specification while  $\varphi \stackrel{\mathcal{N}_{\theta}}\nrightarrow \psi$ means the property specification does not hold on the DNN. 

For a specification set $\{\mathcal{P}_i\}_{i=1}^{p}$ that the DNN $
\mathcal{N}_{\theta}$ should obey, likewise,  $\{\varphi_i\}_{i=1}^{p}\stackrel{\mathcal{N}_{\theta}}\rightarrow \{\psi_i\}_{i=1}^{p}$ and $\{\varphi_i\}_{i=1}^{p} \stackrel{\mathcal{N}_{\theta}}\nrightarrow \{\psi_i\}_{i=1}^{p}$ respectively denotes the satisfiability and unsatisfiability of the specification set on DNN $\mathcal{N}_{\theta}$ and these two relations are defined as
$$\{\varphi_i\}_{i=1}^{p}\stackrel{\mathcal{N}_{\theta}}\rightarrow \{\psi_i\}_{i=1}^{p} \Leftrightarrow \varphi_{i} \stackrel{\mathcal{N}_{\theta}}\rightarrow \psi_{i}, \forall i \in \{1,2,\cdots,p\},$$ 
and
$$\{\varphi_i\}_{i=1}^{p} \stackrel{\mathcal{N}_{\theta}}\nrightarrow \{\psi_i\}_{i=1}^{p}  \Leftrightarrow \varphi_{i} \stackrel{\mathcal{N}_{\theta}}\nrightarrow \psi_{i}, \exists i \in \{1,2,\cdots,p\}.$$ 
That is, a specification set holds on a DNN if and only if each specification in the set holds on the DNN and it does not hold means that at least one specification in the set is violated. The behaviors meeting the relation $\stackrel{\mathcal{N}_{\theta}}\rightarrow$ (resp. $\stackrel{\mathcal{N}_{\theta}}\nrightarrow$) are \emph{expected} (resp. \emph{unexpected}) \emph{behaviors}. Further, the DNNs featuring unexpected behaviors are called \emph{buggy DNNs}.

\section{Problem Formulation}
\label{problem}

Based on aforementioned preliminaries, the \emph{DNN repair problem} comes readily and it is defined as follows.
\begin{problem}(DNN Repair Problem).
Given a buggy DNN $\mathcal{N}_{\theta}$ with respect to a violated  specification set ${\{\mathcal{P}_i\}}_{i=1}^{p}$, i.e., $\{\varphi_i\}_{i=1}^{p} \stackrel{\mathcal{N}_{\theta}}\nrightarrow \{\psi_i\}_{i=1}^{p}$, the DNN repair problem refers to modifying the parameter set $\theta$ to $\theta'$ and the resulting DNN $\mathcal{N}_{\theta'}$ satisfying all the specifications, i.e., $\{\varphi_i\}_{i=1}^{p} \stackrel{\mathcal{N}_{\theta'}}\rightarrow \{\psi_i\}_{i=1}^{p}$.\label{RP}
\end{problem}

Additionally, various DNN repair problems arise depending on the set types, such as point sets, domain sets or others, described by the pre-conditions and post-conditions. Among these, domain sets prove to be more challenging due to the infinity and non-ergodicity of samples within domains, and only several existing repair tools support to patch DNNs over domains {\cite{sotoudeh2021provable}}. Consequently, this paper focuses on repairing DNNs with domain-based specifications and refines Problem \ref{RP} into the following DNN repair problem.
\begin{problem}
(Domain-wise Repair Problem, DRP).
Given a buggy DNN $\mathcal{N}_\theta$ and $\{\varphi_i\}_{i=1}^{p} \stackrel{\mathcal{N}_{\theta}}\nrightarrow \{\psi_i\}_{i=1}^{p}$. For each specification $\mathcal{P}_i$, its pre-condition  $\varphi_i$ and post-conditions $\psi_i$ describe domain sets.  The domain-wise repair problem (DRP) refers to modifying the parameter set $\theta$ to $\theta'$ and forming a new neural network $\mathcal{N}_{\theta'}$ satisfying all the domain-based specifications, i.e., $\{\varphi_i\}_{i=1}^{p} \stackrel{\mathcal{N}_{\theta'}}\rightarrow \{\psi_i\}_{i=1}^{p}$.
\label{DRP}
\end{problem}

For DRPs, input and output domains can be specified as interval boxes, polytopes, and so on. When the behaviors of a DNN violate the property specifications, the regions in the input set that cause the violation are considered \textit{negative input domains} (\textit{negative domains}, for short), denoted by $S_{nd}$. Therefore, domain-wise repair problems, patching a buggy DNN $\mathcal{N}_{\theta}$ with respect to a given specification set $\{\mathcal{P}_i\}_{i=1}^{n}$,  are essentially equivalent to correcting the unexpected behaviours of $\mathcal{N}_{\theta}$ over $S_{nd}$ fundamentally.

Actually, a subtle modification on the DNN parameters may result in a non-negligible impact on the original performance and even introduce extra unexpected behaviors. Consequently, in addition to the requirement of correcting the defective DNNs on certain specification set, it is equally significant for DNN repair approaches to preserve the parameter set or original performance of the buggy DNNs to great extent. This requirement is referred  as the \textit{minimal patch requirement}.

\begin{definition}(Minimal Patch Requirement, MPR). For a DNN repair problem, the generated patch between the original DNN $\mathcal{N}_{\theta}$ and the modified DNN $\mathcal{N}_{\theta'} $ should be minimized, minimizing the impact on the original performance.
\end{definition}

 To this end, some seek to minimize the deviation between $\theta$ and $\theta'$, i.e., minimizing the $\ell$-distance $||\theta'-\theta||_{\ell}$\cite{yang2021reachability,sotoudeh2021provable}. Yet, some pursues  minimizing the performance difference between $\mathcal{N}_{\theta}$ and $\mathcal{N}_{\theta'}$\cite{sun2022causality,sinitsin2020editable,sohn2022arachne}, such as minimizing the accuracy difference $|acc(\mathcal{N}_{\theta'})-acc(\mathcal{N}_{\theta})|$ for classification tasks. MPR is popularly adopted in DNN repair scenarios, and therefore, in what follows, we focus on tackling Problem \ref{DRP} with considering MPR simultaneously.
\section{\bird  \ Repair Framework}
\label{method}

In this section, we give detailed illustrations  on the proposed \bird  \ framework, including the domain behavior description and the behavior-imitation based repair for DRPs, in retraining and fine-tuning styles respectively. 

\subsection{Overall Workflows}

\begin{figure}[tbp]
    \centering
    \includegraphics[scale=0.95]{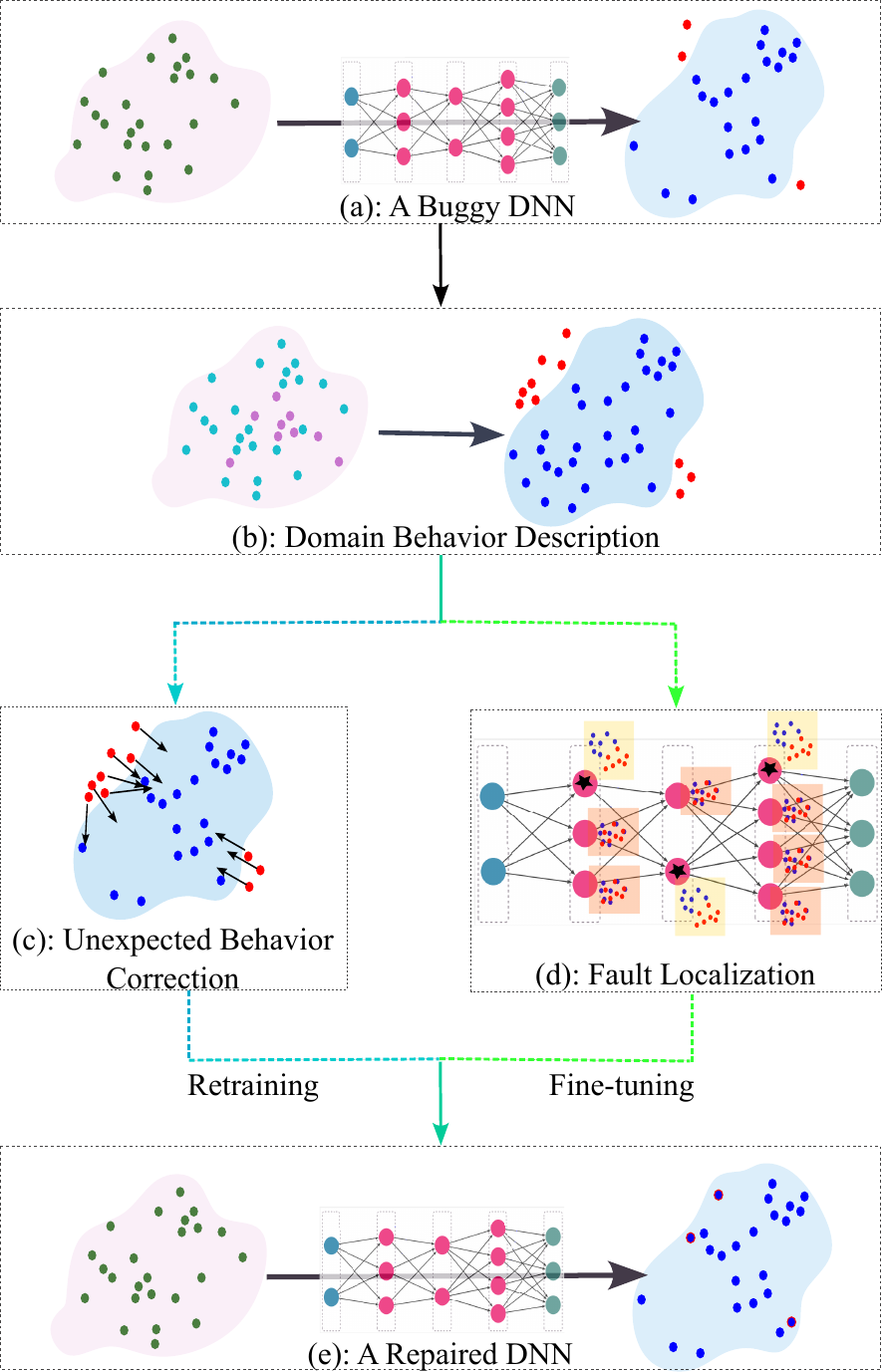}
    \caption{Workflows of \bird \ framework. Retraining workflows: (a)-(b)-(c)-(e); Fine-tuning workflows: (a)-(b)-(d)-(e).}
    \label{fig:workflow}
\end{figure}

The overall workflows of \bird \ are demonstrated in Fig. \ref{fig:workflow}. Beginning with a buggy DNN, its behaviors over negative domains are characterized from the aspects of positive and negative samples. Following it, \bird \ corrects the unexpected predictions of the negative samples towards the nearby expected behaviors during the retraining repair process. Alternatively, \bird \ localizes the most responsible neurons by investigating their behavior differences for the fine-tuning based approach. Subsequently, a repaired DNN is returned from either the retraining based or the fine-tuning based patch.

\subsection{Domain Behavior Description}

First of all, we start with the sample behaviors. Recall the definitions of expected behaviors and unexpected ones, and thus, the input samples w.r.t. the buggy DNN can be correspondingly categorized into \emph{positive samples} and \emph{negative samples.} Similarly, we characterize the buggy DNN's behaviors within an input domain from the perspective of positive and negative samples with Monte Carlo sampling. That is, we collect respective sets of positive and negative samples from the negative domains to illustrate the expected and unexpected behaviors over the negative domains, as shown in Fig.\ref{fig:workflow}b.

To get rid of the infinity and non-ergodicity of samples within domains,  we divide samples within negative domains into the two categories and concentrate on the behaviors of the buggy DNNs on these positive and negative samples when treating the DNN repair problems. The reasons are twofold. On the one hand, the positive and negative predictions generally depict the expected output region and the specification-violated region, depicting the decision boundary of DNNs to some extent. On the other hand, the behaviors of the buggy DNN on the positive samples provide significant guidance for correcting the erroneous behaviors of the original DNNs, i.e., \textit{imitating the expected behaviors of the positive samples}.

A scarce case is that the negative domains absolutely violate the property specification, i.e., we cannot theoretically obtain any positive samples from the negative domains. To overcome this dilemma, we define a relaxed region, termed \emph{domain $\delta$-neighbourhood.}

\begin{definition}(Domain $\delta$-neighbourhood)
For an input domain $D:=[lb_i, ub_i], i\in\{1,2,\cdots,m\}$ and a constant $\delta>0$, where $lb_i$ ($ub_i$) is the lower (upper) bound of the $i$-th input dimension, its domain $\delta$-neighbourhood refers to the region $\delta_{D}:=[lb_i-\delta, ub_i+\delta], i\in\{1,2,\cdots,m\}$\footnote[1]{Here we consider input domains of interval style for example, and the domain $\delta$-neighbourhood of other input domain types can be defined similarly.}.
\end{definition}

The determination of $\delta$ depends on the value that allows the existence of positive samples in the associated domain $\delta$-neighbourhood.   Additionally, it is obvious that on the premise that $\delta$ is a enough small constant, the behaviors of DNNs on the neighborhood samples are similar to the original domain. That is the reason that we collect positive samples in the domain $\delta$-neighbourhood, which is the region slightly relaxed from the negative domains specified by the pre-condition.

Back to the domain behavior characterization, the sample-based characterization provides us a probably approximated correct guarantee on repairing buggy DNNs over negative domains. Assuming a sampling process where the probabilities of positive and negative samples are $q$ and $1-q$, furthermore, assuming a DNN repair process in which a proportion of $u$ of negative samples are repaired while a proportion of $v$ of positive samples disobey the properties, then under a large total sample size, the repaired DNN satisfies the properties with a probability of $q(1-v)+(1-q)u$.

The above-mentioned $u$ and $v$ are termed \emph{property improvement} and \emph{performance drawdown}, catering to repairing buggy DNNs and minimal patch requirement respectively. The Monte Carlo sampling based characterization technique is widely used in statistical model checking\cite{DBLP:series/lncs/LegayLTYSG19,agha2018survey} and there exist some bounds of the total sample number\cite{1959Some,DBLP:conf/icse/LiYHS0Z22}, like Chernoff bound. The sample-based behavior characterization reduces DRPs to sample-based DNN repair problems. Consequently, the main idea of repairing a DNN $\mathcal{N}_{\theta}$ with respect to a negative input domain set $S_{nd}$ (Problem \ref{DRP}) is to correct the unexpected behaviors of negative samples on the negative domains.  In what follows, we present our repair approaches to DRPs, in aspects of retraining and fine-tuning styles.
 
\subsection{Retraining Approach to DRP}
\label{Sec:re}

Overall, the retraining style patch of DRPs is composed of three procedures, that is, \textsc{sampleCollect}, \textsc{negativeCorrect} and \textsc{retrainDNN}, as shown in Alg.\ref{alg: retraining for drp}. The positive and negative samples within the negative domains are collected in \textsc{sampleCollect},  together with their output behaviors, and \textsc{negativeCorrect} (i.e., Alg. \ref{alg: nbc}), subsequently, corrects the unexpected behaviors and prepares the repairing dataset via imitating the expected behaviors of the positive samples. Finally, the procedure \textsc{retrainDNN} retrains the original DNN, pursuing eliminating unexpected behaviors and keeping minimal performance drawdown synergistically.

The procedure \textsc{sampleCollect} is shown in Line  \ref{ndc_start}-\ref{ndc_end} in Alg. \ref{alg: retraining for drp}. We collect sample sets ${S}_{p}$ and $S_n$, composed of positive and negative samples from the negative domains extracted from the violated properties. It is notable that maybe a domain $\delta$-neighbourhood is needed, and the $\delta$ value should not only be as small as possible,  but ensure a required number of positive samples being sampled. In the meanwhile, we also record their output behaviors $Y_p$ and $Y_n$ concerning the buggy DNN for the subsequent negative behavior correction.

\begin{algorithm}[tb]
\caption{Retraining Patch for DRPs}
\label{alg: retraining for drp}
\textbf{Input}: a buggy DNN $\mathcal{N}_\theta$, a specification set $\{\mathcal{P}_i\}_{i=1}^{p}$, training dataset $\mathcal{D}$.\\
\textbf{Output}: a modified DNN $\mathcal{N}_{\theta'}$.

\begin{algorithmic}[1]

\Procedure{sampleCollect}{$\mathcal{N}_\theta$, $\{\mathcal{P}_i\}_{i=1}^{p}$} \label{ndc_start}
    \State extract negative domain set $S_{nd}$ with $\mathcal{N}_\theta$ and $\{\mathcal{P}_i\}_{i=1}^{p}$
      \State  $\mathcal{D}_{re} \leftarrow \emptyset$
     \For{\textbf{each} $s$ \textbf{in} $S_{nd}$}
        \State $S_p, S_n \leftarrow$ sets of positive and negative samples
        \State $Y_p \leftarrow \{\mathcal{N}_{\theta}(\bm{x}_{p})\}_{i=1}^{\#S_{p}},  \bm{x}_{p} \in S_{p}$ 
        \State $Y_n \leftarrow \{\mathcal{N}_{\theta}(\bm{x}_{n})\}_{i=1}^{\#S_{n}},  \bm{x}_{n} \in S_{n}$ \label{ndc_end}
        \State $Y_c\leftarrow$ \textsc{negativeCorrect}($Y_p, Y_n$) \label{call:nbc}
        \State $\mathcal{D}_{re} \leftarrow \mathcal{D}_{re} \cup \{\{S_p\cup S_n\}, \{Y_p \cup Y_c\}\}$ \label{repair_data}
     \EndFor
     \State \textbf{return} $\mathcal{D}_{re}$

\EndProcedure \label{dpr_end}

\Procedure{retrainDNN}{$\mathcal{N}_{\theta}$, $\mathcal{D}$, $\mathcal{D}_{re}$} \label{rs_start}
\State $\mathcal{N}_{\theta'} \leftarrow$ train $\mathcal{N}_{\theta}$ with $\mathcal{D}$ and $\mathcal{D}_{re}$ 
\State \textbf{return} $\mathcal{N}_{\theta'}$
\EndProcedure \label{rs_end}

\end{algorithmic}

\end{algorithm}

 Alg. \ref{alg: nbc} demonstrate the procedure \textsc{negativeCorrect} detailedly (Line \ref{call:nbc} in Alg. \ref{alg: retraining for drp}). Due to the characteristic of the DNN training process, i.e., the loss functions are defined on the output layer. Here, we only focus on the final part of the DNN behaviors, the predictions (i.e., $\mathcal{N}_{\theta}(\cdot)$ or $\mathcal{N}_{\theta}^{L-1}(\cdot)$), and eliminate the negative behaviors via imitating their surrounding positive behaviors. For each negative prediction $\bm{y}_n \in Y_n$, the top $k$ closest positive predictions to  it are selected and their mean value is assigned as the new label for the corresponding negative sample if the mean value satisfies the post-condition that $\bm{y}_n$ violates. Or, the closest positive prediction is chosen as the candidate label for the following DNN retaining. Obviously, the presented label assignment ensures the correctness of the newly assigned label for each negative input sample. With the calibrated labels, the repairing dataset $\mathcal{D}_{re}$ is done in Line \ref{repair_data} in Alg. \ref{alg: retraining for drp}. This statement is visualized in Fig. \ref{fig:workflow}c.

When comes to the final procedure \textsc{retrainNN} with training dataset $\mathcal{D}$ and repairing dataset $\mathcal{D}_{re}$,  Line  \ref{rs_start}-\ref{rs_end} of Alg. \ref{alg: retraining for drp}, it is of vital importance to clarify the formulation of \textit{loss function} utilized for the retraining process.

\begin{algorithm}[tbp]
\caption{Negative Behavior Correction}
\label{alg: nbc}
\textbf{Input}: positive predictions $Y_p$, negative predictions $Y_n$, neighbor number $k$.\\
\textbf{Output}: corrected negative predictions $Y_c$.

\begin{algorithmic}[1]

\Procedure{negativeCorrect}{$Y_p$, $Y_n$} \label{nbc_start}
    \State $Y_c \leftarrow \emptyset$
     \For{\textbf{each} $\bm{y}_n$ \textbf{in} $Y_n$}
        \State $\bm{d} \leftarrow$ distances between $\bm{y}_n$ and $\bm{y}_p$, $\bm{y}_p \in Y_p$
        \State $\bm{d} \leftarrow $ sort $\bm{d}$ in an increasing order
        \State $\bm{y}_c \leftarrow $ the average value of the first $k$ items in $\bm{d}$
        \If {\textit{$\bm{y}_c$ is an expected behavior}} 
        \State $Y_c \leftarrow Y_c \cup \{\bm{y}_c\}$
         \Else \State $Y_c \leftarrow Y_c \cup \{\bm{d}_0\}$ {\Comment{The first item in $\bm{d}$.}}
        \EndIf 
     \EndFor
     \State \textbf{return} $Y_c$
     
\EndProcedure \label{nbc_end}

\end{algorithmic}

\end{algorithm}

\subsubsection{Loss formulation for DRPs} The repairing dataset $\mathcal{D}_{re}:=\{\bm{x}_{re}^{i}, \bm{l}_{re}^{i}\}_{i=1}^{\#\mathcal{D}_{re}}$ is constructed to eliminate the unexpected behaviors on the negative domain set ${S}_{nd}$.  The retraining process aims to minimize the discrepancy between the $\mathcal{N}_{\theta}$'s predictions and the corrected labels, which is formulated in Eqn. \eqref{Lsrp}
\begin{equation}
    \mathcal{L}_\text{DRP}(\theta'):=\sum_{i=1}^{\#\mathcal{D}_{re}}||\mathcal{N}_{\theta'}(\bm{x}^{i}_{re})-\bm{l}_{re}^{i}||_{\ell},
    \label{Lsrp}
\end{equation}
where $||\cdot||_{\ell}$ is the $\ell$-norm distance. It can be seen that the unexpected behaviors on the negative samples are thoroughly eliminated under the condition that the loss function defined in Eqn. \eqref{Lsrp} is minimized to 0 during the retraining process, which is called provable DNN repair\cite{sotoudeh2021provable}.

\subsubsection{Loss formulation for MPR} Different from traditional DNN training, MPR  requires the modified DNN $\mathcal{N}_{\theta'}$ to preserve the original performance of $\mathcal{N}_{\theta}$. Consequently, we also take the original input-output dataset $\mathcal{D}:=\{(\bm{x}^{i}, \bm{l}^{i})\}_{i=1}^{\#\mathcal{D}}$ into account and pursue the minimal performance drawdown by loss function $\mathcal{L}_\text{MPR}(\theta')$, which is shown in Eqn.\eqref{Lmpr}.
\begin{equation}
     \mathcal{L}_\text{MPR}(\theta'):=\sum_{i=1}^{\#\mathcal{D}}||\mathcal{N}_{\theta'}(\bm{x}^{i})-\bm{l}^{i}||_{\ell}
     \label{Lmpr}
\end{equation}

Therefore, for procedure \textsc{retrainDNN}, dealing DRPs with MPR is a multi-objectives optimization and the final loss function can be formulated as 
\begin{equation}
    \underset{\theta'}{\text{minimize}}
  (\alpha\cdot\mathcal{L}_\text{DRP}(\theta') + \beta\cdot\mathcal{L}_\text{MPR}(\theta'))
\end{equation}
where $\alpha,\ \beta \in [0,1]$, $\alpha+\beta=1$ and $\theta'$ is initialized as $\theta$. The retraining process patches the original DNN $\mathcal{N}_{\theta}$ without considering MPR when the configuration is $\alpha=1,\ \beta=0$ and it only focuses on minimal parameter change with the configuration $\alpha=0, \ \beta=1$. The procedure \textsc{retrainDNN} terminates if the negative samples meet the specifications or the maximum iteration number is reached, and a repaired DNN $\mathcal{N}_{\theta'}$ is returned in Line \ref{rs_end} of Alg. \ref{alg: retraining for drp}, as shown in Fig. \ref{fig:workflow}e.

\subsection{Fine-tuning Approach to DRPs.} 

\begin{algorithm}[tbp]
\caption{Fine-tuning Patch for DRPs}
\label{alg: fine tuning for drp}
\textbf{Input}: a buggy DNN $\mathcal{N}_\theta$, a specification set $\{\mathcal{P}_i\}_{i=1}^{p}$.\\
\textbf{Output}: a modified DNN $\mathcal{N}_{\theta'}$.

\begin{algorithmic}[1]

\Procedure{sampleCollect}{$\mathcal{N}_\theta$, $\{\mathcal{P}_i\}_{i=1}^{p}$} \label{ndc_start2}
    \State extract negative domain set $S_{nd}$ with $\mathcal{N}_\theta$ and $\{\mathcal{P}_i\}_{i=1}^{p}$
     
     \For{\textbf{each} $s$ \textbf{in} $S_{nd}$}
        \State $S_p, S_n \leftarrow$ sets of positive and negative samples \label{ndc_end2}
        \State $\bm{R}\leftarrow \{\bm{r}^{1},\cdots, \bm{r}^{L-1}\}$, where $\bm{r}^{i}\leftarrow \bm{0}\in \mathbb{R}^{d_i}$
        \State $\bm{R}' =$ \textsc{behaveAnalyze}($\mathcal{N}_{\theta},S_p, S_n, \bm{R}$)  \label{ba}
        \State $\bm{R} = \bm{R} + \bm{R}'$ {\Comment{Element-wise addition.}}
    \EndFor
    \State \textbf{return} $\bm{R}$
    \EndProcedure
    
\Procedure{fineTuneDNN}{$\mathcal{N}_{\theta}$, $\bm{R}$} \label{ftnn_start}
\State sort $\bm{R}^{ij}, i\in\{1,\cdots,L-1\}$, $j\in\{1,\cdots,d_i\}$
\State $I \leftarrow$ indices of the largest $r$ items in $\bm{R}$
\State ${\theta'} \leftarrow$ modified  weights of neurons in $I$  \label{PSOft}
\State \textbf{return} $\mathcal{N}_{\theta'}$ \label{returnmn}
\EndProcedure \label{ftnn_end}
\end{algorithmic}

\end{algorithm}

Except for the retraining based repair for DRPs, \bird \ also provides an approach of fine-tuning style patch for DRPs. The algorithm is demonstrated in Alg. \ref{alg: fine tuning for drp}. Likewise, the algorithm can be divided into three procedures, i.e., \textsc{sampleCollect}, \textsc{behaveAnalyze} and \textsc{fineTuneDNN}. After collecting the sets of positive and negative samples in procedure \textsc{sampleCollect} (Line \ref{ndc_start2}-\ref{ndc_end2}), procedure  \textsc{behaveAnalyze} (Line \ref{ba}) completes the fault localization by analysing the discrepancies between the behaviors on the negative samples (unexpected behaviors) and those on the positive  samples (expected behaviors).  Then, \textsc{fineTuneDNN} (Line \ref{ftnn_start}-\ref{ftnn_end}) tunes the weights of most responsible neurons for the unexpected behaviors with optimization methods.

Compared with only focusing on the final predictions in Section \ref{Sec:re}, \bird \ makes a more insightful investigation on neuron behaviors in procedure \textsc{behaveAnalyze}, which is illustrated in  Alg. \ref{alg:response}. Recall the definition of behaviors, the behaviors of DNN $\mathcal{N}_{\theta}$ on a negative input sample $\bm{x}_n$ and the positive sample set $S_{p}$ are respectively denoted by 
$$\{\bm{x}_{n}, \mathcal{N}^{1}_{\theta}(\bm{x}_{n}),  \mathcal{N}^{2}_{\theta}(\bm{x}_{n}),\cdots,  \mathcal{N}^{L-1}_{\theta}(\bm{x}_{n})\}$$ and  $$\{\bm{x}_p^{j}, \mathcal{N}^{1}_{\theta}(\bm{x}_p^{j}),  \mathcal{N}^{2}_{\theta}(\bm{x}_p^{j}),\cdots,  \mathcal{N}^{L-1}_{\theta}(\bm{x}_p^{j})\}_{j=1}^{\#S_{p}}.$$
 Subsequently, the \textit{layer responsibility} $\bm{r}^{i}$ of the neurons  on the $i$-th layer for the unexpected behaviors are defined as follows,
\begin{equation}
    \bm{r}^{i}:=\sum_{j=1}^{{\#S_{p}}}| \mathcal{N}_{\theta}^{i}(\bm{x}_{n})-\mathcal{N}_{\theta}^{i}(\bm{x}_{p}^{j})|,\  i \in \{1,2,\cdots, L-1\}.
    \label{response_init}
\end{equation}
Further, the \emph{neuron responsibility} $\bm{r}^{ij} \ (0\le j<d_{i})$ indicates to what extent, the $j$-th neuron on the $i$-th layer is responsible for the unexpected behaviors, and we name $\bm{R}:=\{\bm{r}^{i}\}_{i=1}^{L-1}$ the \emph{responsibility matrix}. Essentially, the main idea of fault localization is also behavior imitation. The responsibility matrix reveals the behavior difference of each neuron over negative and positive samples and the larger a neuron's behavior difference, the more necessary for it to imitate expected behaviors to minimize the discrepancy.  Since there is no need to change the input layer when repairing a buggy DNN, and thus it is unnecessary to define the responsibility metric for the input neurons\footnote{Actually, the defined responsibility metric also works for the input neurons.}. Likewise, these statements are visualized in Fig. \ref{fig:workflow}d.

\begin{algorithm}[tb]
\caption{Fault Localization of Responsible Neurons}
\label{alg:response}
\textbf{Input}: a buggy DNN $\mathcal{N}_\theta$, positive sample set $S_p$, negative sample set $S_n$, initialized responsibility matrix $\bm{R}$.\\
\textbf{Output}: computed responsibility matrix $\bm{R}$.

\begin{algorithmic}[1]

\Procedure{{behaveAnalyze}}{$\mathcal{N}_{\theta},S_p, S_n, \bm{R}$} \label{bas_start}
\For {$i\leftarrow1$ \textbf{to} $L-1$}  
\For{\textbf{each} $\bm{x}_n$ \textbf{in} $S_n$} \label{for1}
\For {\textbf{each} $\bm{x}_p$ \textbf{in} $S_p$} 
\State $\bm{R}^{i}\leftarrow \bm{R}^{i} + |\mathcal{N}^{i}_{\theta}(\bm{x}_{n})-\mathcal{N}^{i}_{\theta}(\bm{x}_{p})|$\label{for2}
\EndFor
\EndFor
\EndFor
\State \textbf{return} $\bm{R}$
\EndProcedure \label{bas_end}
\end{algorithmic}
\end{algorithm}

 In addition, we also modify Eqn. \eqref{response_init} to Eqn. \eqref{response_modify} to break out the two loop in Alg. \ref{alg:response} (Line \ref{for1}-\ref{for2}), improving the efficiency of the fault localization further.
\begin{equation}
    \bm{r}^{i}:=|\sum_{t=1}^{{\#S_{n}}} \mathcal{N}_{\theta}^{i}(\bm{x}_{n})-\sum_{j=1}^{{\#S_{p}}}\mathcal{N}_{\theta}^{i}(\bm{x}_{p}^{j})|,\  i \in \{1,\cdots, L-1\}.
    \label{response_modify}
\end{equation}

After the computation of the responsibility matrix, procedure \textsc{fineTuneDNN} (Line \ref{ftnn_start}-\ref{ftnn_end} in Alg. \ref{alg: fine tuning for drp}) sorts the responsibility matrix and selects the  most responsible (largest) $r$ neurons, whose weights are to be optimized subsequently. As for the optimization method, we choose Particle Swarm Optimization (PSO)\cite{kennedy1995particle} for its convergence speed and accuracy in continuous optimization, and other algorithms are also alternative, like Differential Evolution (DE)\cite{storn1997differential}, Markov Chain Monte Carlo (MCMC)\cite{fok2017optimization} and Pelican Optimization Algorithm\cite{trojovsky2022pelican}. 

Considering multiple particles in the search space, $\Vec{x_i}$ and $\Vec{v_i}$ stand for the location and velocity of each particle. During each search iteration, PSO updates $\Vec{x_i}$ and $\Vec{v_i}$ according to a fitness function. The evolution rule update $\Vec{v_i}$ based on current velocity $\Vec{v_i}$, the best local location previous found $\Vec{p_i}$ and the best global location previous found $\Vec{p_g}$ and it update $\Vec{x_i}$ with current velocity $\Vec{v_i}$ and location $\Vec{x_i}$ then. The PSO update equation is formulated as follows\cite{shi1998parameter}.
\begin{equation}
\begin{aligned}
& \Vec{v_i}\leftarrow \omega\Vec{v_i} + R(0,c_1)(\Vec{p_i}-\Vec{x_i}) + R(0, c_2)(\Vec{p_g}-\Vec{x_i}), \\
& \Vec{x_i}\leftarrow \Vec{v_i} + \Vec{v_i},
\end{aligned}
\end{equation}
where $\omega,\ c_1, \ c_2$ respectively represent the inertia weight, cognitive parameter and social parameter. $R(0,c)$ refers to a random value sampled from $[0,c]$. More importantly, the fitness function in PSO determines the best location.

Come back to Line \ref{PSOft} of procedure \textsc{fineTuneDNN}, the weights of the  most responsible $r$ neurons are the ``particles'' for PSO and their locations are initialized as the original value in $\theta$, their velocities are set as zeros. The fitness function $\mathcal{F}$ for parameter optimization is defined following.
\begin{equation}
    \mathcal{F}:=\alpha \cdot \text{unexpBeh} + \beta \cdot \text{drawDown}
    \label{fit func}
\end{equation}
where $\alpha,\ \beta \in [0,1]$ and $\alpha+\beta=1$. \emph{unexpBeh} (unexpected behaviors) indicates the percentage of the negative samples violating the specifications and \emph{drawdown} is the DNN performance loss, such as the dropout of classification accuracy. The construction of Eqn. \eqref{fit func} means the PSO attempts to minimize both the unexpected behaviors and the performance drawdown.  Moreover, the algorithm terminates if there generates an intolerable performance  degeneration between the original and modified DNNs, corresponding to MPR,  or the maximum search number is reached and returns the repaired DNN $\mathcal{N}_{\theta'}$ in Line \ref{returnmn} of Alg. \ref{alg: fine tuning for drp}, as shown in Fig. \ref{fig:workflow}e.

\section{Experiments}
\label{exp}
\subsection{Experiment Setup} 
In this section, the proposed  \bird \ framework is evaluated and compared with some state-of-the-art DNN repair works on the ACAS Xu DNN repair problem, in aspects of retraining and fine-tuning styles, which are respectively termed \emph{\bird-RT} and \emph{\bird-FT} hereafter.

\subsubsection{ACAS Xu DNN repair problem} The ACAS Xu DNN repair problem, patching ACAS Xu DNNs \cite{DBLP:journals/corr/abs-1810-04240} with respect to certain safety properties, is a widely utilized and evaluated DNN domain-wise repair problem. ACAS Xu DNNs contains an array of 45 DNNs (organized as a $5 \times 9$ array) that produce horizontal maneuver advisories of the unmanned version of Airborne Collision Avoidance System X, which is a highly safety-critical system developed by the Federal Aviation Administration. More concretely, these DNNs take a five-dimensional input, representing the scenario around the aircraft, and output a five-dimensional prediction, indicating five possible advisories. The ACAS Xu DNNs are associated with 10 safety properties, $\phi_1$-$\phi_{10}$, and their pre-/post-conditions are described as polytope sets, satisfying the definition of DRPs. These safety properties require that the output corresponding to an input within a specific input region must fall in the given safe polytope(s). Some verification works \cite{DBLP:conf/cav/KatzBDJK17,DBLP:journals/corr/abs-2007-11206} discovery that among these 45 DNNs, there are 35 DNNs violate at least one safety property, e.g., $N_{2,9}$ violates the safety property $\phi_8$. 

We follow the ACAS Xu DNN evaluation cases adopted in previous work Veritex\cite{yang2022neural}, CARE \cite{sun2022causality} and PRDNN\cite{sotoudeh2021provable}. Herein, as classified in Veritex, we select one simple test case $N_{3,3}$, and two hard test cases, $N_{1,9}$ and $N_{2,9}$ to evaluate the \bird \ framework. The detailed information of these used DNNs is displayed in Table \ref{tab:NN_info} (\textit{DNN}$_{1}$, \textit{DNN}$_{2}$ and \textit{DNN}$_{3}$). 

\begin{table}[htbp]
    \centering
    \caption{DNN models used in Experiments.}
    \label{tab:NN_info}
    \setlength{\tabcolsep}{1mm}{
    \begin{tabular}{cccc}
      \Xhline{1.2pt}
        DNN   & Description    & Architecture  & \#Param  \\ 
        \hline
        \textit{DNN{$_1$}} & ACAS Xu, $N_{3,3}$ & 7-layer Relu FFNN & 13,350\\
       \textit{DNN{$_2$}} & ACAS Xu, $N_{1,9}$ & 7-layer Relu FFNN & 13,350 \\
        \textit{DNN{$_3$}} & ACAS Xu, $N_{2,9}$ & 7-layer Relu FFNN & 13,350 \\
         \textit{DNN{$_4$}} & Modified ACAS Xu, $N_{1,9}$ & 7-layer Tanh FFNN & 13,350 \\
          \textit{DNN{$_5$}} &  Modified ACAS Xu, $N_{1,9}$ & 7-layer LeakyReLU FFNN & 13,350 \\
         \textit{DNN{$_6$}} &  Modified ACAS Xu, $N_{1,9}$ & 7-layer ELU FFNN & 13,350 \\
    \Xhline{1.2pt}
    \end{tabular}}
    
\end{table}

\subsubsection{Evaluation platform} All the experiments are conducted on a machine with 12th Gen Intel(R) Core(TM) i9-12900H 2.50 GHz and 16 GB system memory.
The codes to reproduce our experimental results are available at \url{https://github.com/ByteTao5/BIRDNN}.

\subsubsection{Research questions}
We report experimental results to answer the following  research questions, demonstrating \bird's effectiveness, efficiency and compatibility:
\begin{itemize}
    \item \emph{\textbf{RQ1:} Can \bird-RT repair the buggy DNNs successfully? How about its efficiency compared to other retraining  based methods?}
    \item \emph{\textbf{RQ2:} Can \bird-FT repair the buggy DNNs successfully? How about its efficiency compared to other fine-tuning based methods?}
    \item \emph{\textbf{RQ3:} What about the compatibility of \bird \ on DNNs with different activation functions?}
\end{itemize}

\subsection{Evaluations on \bird-RT} 
The retraining performance of \bird-RT on ACAS Xu DNNs is compared with Veritex, a provable retraining repair method. Notably, the original training data of ACAS Xu DNNs are not publicly available online. To make fair comparisons, we follow the experiment settings adopted in Veritex, and we uniformly sample a set of 10K training data and 5K test data from the input space. Veritex does not distinguish the positive and negative samples in the training data explicitly in advance, and the violated safety properties are improved by minimizing the distance between the unsafe output region (exact polytope regions by reachability analysis) and the safety region. In contrast, \bird \ requires reassigning the negative samples of the training data with the predictions of the nearest positive samples, and we respectively set the proportions of the negative samples and positive samples as 10\% and 90\% in what follows. Additionally, the weights balancing the original performance and safety improvement are set as $\alpha=\beta=0.5$. We also introduce an early stop mechanism that the retraining process terminates when the safety improvement keeps stable within 10 successive epochs. The early stop also ensures minimal patch requirement to some extent.

\begin{table}[htbp]
    \centering
    \caption{Performance Comparisons between \bird-RT and  Veritex.}
    \label{tab:DRP_retrain}
    \setlength{\tabcolsep}{1mm}{
    \begin{tabular}{ccccc}
      \Xhline{1.2pt}
        \multicolumn{1}{c}{\multirow{2}{*}{DNN}} & \multicolumn{2}{c}{Veritex} & \multicolumn{2}{c}{\bird-RT} \\ \cline{2-5}
     & Accuracy    & Time  & Accuracy    & Time \\
     \hline

   \textit{DNN}$_1$ & 100\% &  4.96 &  100\%   & \textbf{ 1.3}    \\ 
    \textit{DNN}$_2$ & 100\% &  Timeout$^{*}$ (11250.7) &  100\%   &  \textbf{174.70}   \\ 
    \textit{DNN}$_3$ & 100\% & 2167.16 &  100\%   &  \textbf{87.45}   \\ 

     \Xhline{1.2pt}
     \multicolumn{2}{l}{\emph{literature reported result$^{\dag}$}:} &  & \\
     Time  &   \multicolumn{2}{c}{Accuracy} & {\textit{DNN}$_3$} & {\textit{DNN}$_4$} \\
     \hline
    ART &  \multicolumn{2}{c} {89.08\% $ \sim $  98.06\%}  & 67.5 & 72.6\\
    ART-refinement & \multicolumn{2}{c} {88.82\% $ \sim $  95.85\%} & 82.5  & 88.4\\
    \Xhline{1.2pt}
    
    \end{tabular}}
        \begin{tablenotes}    
        \footnotesize               
        \item \textbf{*} The running time limit is set as 3 hours. Here we provide the running time reported in Veritex\cite{yang2022neural}  in the parentheses.
       \item \textbf{$\dag$} The listed results are recorded in Veritex\cite{yang2022neural}.
      \end{tablenotes}
\end{table}

\textbf{\emph{Repair performance.}} The performance of \bird-RT on ACAS Xu DNNs is displayed in Table \ref{tab:DRP_retrain}, versus the state-of-the-art work, Veritex, in terms of the \emph{Accuracy} on the test data and the \emph{Running Time}. Because the test set contains both positive and negative samples, the \emph{Accuracy} indicator combines safety improvement and performance drawdown, and when the accuracy indicator reaches 100\%, it means complete repair and no performance loss. Moreover, we also list the performance of ART\cite{DBLP:conf/fmcad/LinZSJ20}, a DNN property-guided  training method, for a more comprehensive reference. It can be observed that both Veritex and \bird-RT repair these buggy DNNs successfully, i.e., with an accuracy of 100\% in the test data. As for the running time, Veritex takes a large amount of running time, dominantly by the computation of exact output regions, while \bird-RT significantly reduces the time consumption occupied by Veritex, ranging from 73.8\% (\textit{DNN}$_{1}$),  95.96\%   (\textit{DNN}$_{3}$) to 98.45\% (\textit{DNN}$_{2}$). Meanwhile, compared with ART (or ART-refinement) method, which starts from scratch to train safe ACAS Xu DNNs, \bird-RT takes a little more time, while reaching higher accuracy.
\begin{framed}
     \noindent \textit{\textbf{Answer RQ1:} \bird-RT can successfully repair the buggy DNNs, almost without no performance loss, and it provides much more efficient solutions to DRPs, compared with state-of-the-art works.} 
\end{framed}

\subsection{Evaluations on \bird-FT}
\label{evl-ft}
As for patching ACAS Xu DNNs with fine-tuning based method, in this subsection, we mainly evaluate the performance of \bird-FT from the aspects of \emph{cross-layer repair} and \emph{layer-wise repair}. The former allows us to localize responsible neurons from multiple layers, while the latter restricts the neurons to be tuned must locate on the same layer. We compare \bird-FT with the state-of-the-art tools, CARE and PRDNN, respectively corresponding to the cross-layer repair and layer-wise repair.

\subsubsection{Cross-layer Repair Performance} 
Likewise, we utilize the same setups utilized by CARE herein. For each DNN, a set of 10K negative data are randomly sampled to improve the safety property, and another set of 10K positive samples are randomly collected to evaluate the drawdown of the original performance. Another two sets of the same settings are collected for the testing.  We set the number of the neurons to be repaired as 10. During the running of the PSO algorithm, as generally recommended in \cite{DBLP:journals/swarm/PoliKB07},  the parameters are set as $\omega=0.8, c_1=c_2=0.41$, and the particle number and the maximum iteration number are 20 and 100, respectively. 
Moreover, to reduce the searching time, the search terminates if it fails to find a better optimization location in 10 consecutive iterations. The weight configuration of balancing performance drawdown and safety improvement is $\alpha=0.6, \beta=0.4$.

\textbf{\emph{Repair Performance.}} The performance on the test data of \bird-FT and CARE are compared in Table \ref{tab:DRP_FT}, in aspects of \emph{Improvement}, \emph{Drawdown}, \emph{Localization Time} (\emph{Loc Time}), and \emph{Total Time}. Improvement and Drawdown indicate the repair performance on safety correction and accuracy maintenance, respectively. Localization Time and Total Time are the running time consumed by the fault localization and the whole patching procedures.

\begin{table}[htbp]
    \centering
    \caption{Performance Comparisons of \bird-FT and CARE.}
    \label{tab:DRP_FT}
    \setlength{\tabcolsep}{3mm}{
    \begin{tabular}{ccccc}
      \Xhline{1.2pt}
      Method & Item  &   \textit{DNN}$_1$ &  \textit{DNN}$_2$  &  \textit{DNN}$_3$  \\ 
      \hline
      \multicolumn{1}{c}{\multirow{4}{*}{CARE}} & Improvement    &  1.0 &   0.9999  & 0.9935 \\
     & Drawdown    & 0.0  &  0.0   & 0.0 \\
     & Loc Time    &  35.96 &   41.33  & 36.24 \\
     & Total Time    & 201.32  &   370.40  &  197.55\\
     \hline

     \multicolumn{1}{c}{\multirow{4}{*}{\bird-FT}} & Improvement    & 1.0  & \textbf{1.0}    & \textbf{1.0} \\
     & Drawdown    &  0.0 &  0.0   & 0.0 \\
     & Loc Time    &  \textbf{0.067} &   \textbf{0.071}  & \textbf{0.08} \\
     & Total Time    & \textbf{16.80}  &   \textbf{16.74}  & \textbf{15.95} \\
     \Xhline{1.2pt}
    \end{tabular}}
\end{table}

From Table \ref{tab:DRP_FT}, it can be seen that \bird-FT repairs all the buggy DNNs successfully without performance drawdown, which is a little better than CARE. CARE provides almost complete repair on these DNNs and its patch also loses no original accuracy. As for the running time, our proposed \bird-FT greatly reduces the time consumption, both the localization time and total time. The behavior-imitation based fault localization of \bird-FT is much more efficient than the causality-based one of CARE, significantly reducing about 99.8\% computation time. According to the final repair performance versus CARE, however, the proposed lightweight fault localization does not mean the scarification of safety improvement or original performance. Due to the same optimization algorithm (PSO), by contrast, \bird-FT finds a more appropriate set of responsible neurons resorting to behavior analysis, and it reduces the following search time in the PSO algorithm further, which only takes about 4\%$\sim$8\% of the total time taken by CARE.

\subsubsection{Layer-wise Repair Performance} When we consider the responsibility matrix row by row, it provides us an indication of which neurons located on a certain layer are necessary to repairing, i.e., we select responsible neurons only in a given layer for repair, named \emph{layer-wise DNN repair}. PRDNN is one of the most representative layer-wise repair works. It is a provable fine-tuning repair method based on the DNN verification techniques with the polytope domain, and it repairs a specific layer of the buggy DNNs without fault localization. The layer-wise repair performance of \bird-FT and PRDNN on \textit{DNN}$_{3}$ are displayed in Table \ref{tab:DRP_FT_layer}. It can be observed that PRDNN only succeeds in repairing the buggy DNN on the output layer and fails on other layers. It results from the program hangs caused by dimensionality curse encountered in the computation of exact linear regions. By contrast, \bird-FT works successfully for every layer and can reach almost complete repair (except \textit{L}$_4$) without performance drawdown. Likewise, as for the running time,  \bird-FT is more efficient than PRDNN.

\begin{table}[htbp]
    \centering
    \caption{Layer-wise Repair Performance  Comparisons of \bird-FT and PRDNN.}
    \label{tab:DRP_FT_layer}
  \setlength{\tabcolsep}{3mm}{
    \begin{tabular}{ccccc}
      \Xhline{1.2pt}
        \multicolumn{1}{c}{\multirow{2}{*}{Layer}} & \multicolumn{2}{c}{PRDNN} & \multicolumn{2}{c}{\bird-FT} \\ \cline{2-5}
     & Improve/Acc  & Time  & Improve/Acc    & Time \\
     \hline

   \textit{L}$_1$ & --$^{*}$ &  -- &  \textbf{100\%/100\%}   & \textbf{ 17.83}    \\ 
    \textit{L}$_2$ & -- &  -- &  \textbf{100\%/100\%}   &  \textbf{17.30}   \\ 
    \textit{L}$_3$ & -- & -- &  \textbf{100\%/100\% } &  \textbf{17.19}   \\ 
    \textit{L}$_4$ & -- &  -- &  \textbf{99.96\%/100\%}   & \textbf{ 17.08}    \\ 
    \textit{L}$_5$ & -- &  -- &  \textbf{100\%/100\%}  &  \textbf{17.80}   \\ 
    \textit{L}$_6$ & 100\%/100\% & 20.50 &  100\%/100\%   &  \textbf{19.57}   \\ 
     \Xhline{1.2pt}
    \end{tabular}}
      \begin{tablenotes}    
        \footnotesize               
        \item \textbf{*} PRDNN fails to repair the layers except for the final (output) layer.
      \end{tablenotes}
\end{table}

Moreover, we list the average improvement (\emph{Ave\_Improve}) and average drawdown (\emph{Ave\_Drawdown}) of \bird-FT, CARE and PRDNN in Table \ref{tab:DRP_FT_ave} on the three DNNs.  In particular, PRDNN fails to repair \textit{DNN}$_2$ due to program hangs and the average results are evaluated on repairing \textit{DNN}$_1$ and \textit{DNN}$_3$. As demonstrated in Table \ref{tab:DRP_FT_ave},  \bird-FT outperforms the previous methods, CARE and PRDNN, repairing the buggy DNNs completely and with no performance drawdown.

\begin{table}[htbp]
    \centering
    \caption{Average Performance of \bird-FT , CARE and PRDNN.}
    \label{tab:DRP_FT_ave}
    \setlength{\tabcolsep}{4mm}{
    \begin{tabular}{cccc}
      \Xhline{1.2pt}
      &   CARE &  PRDNN  &  \bird-FT  \\ 
      \hline
    Ave\_Improve & 99.6\%    &  97.4\% &  \textbf{100\%}    \\
    Ave\_Drawdown & 0.15\%   &  0.01\% &  \textbf{0\%}  \\
     \Xhline{1.2pt}
    \end{tabular}}
\end{table}

\subsubsection{Hyperparameter Comparisons} Additionally, we make some comparisons on the performance of different hyperparameter selections. We change the balancing weight $\alpha$ (then $\beta=1-\alpha$)  to evaluate the cross-layer repair performance of \bird-FT on \textit{DNN}$_{2}$ and set different repaired neuron numbers for the layer-wise repair on \textit{DNN}$_{3}$.

\begin{figure}[htbp]
    \centering
    \subfloat[Comparisons on different $\alpha$ values.]{\label{alpha}\includegraphics[ scale=0.54]{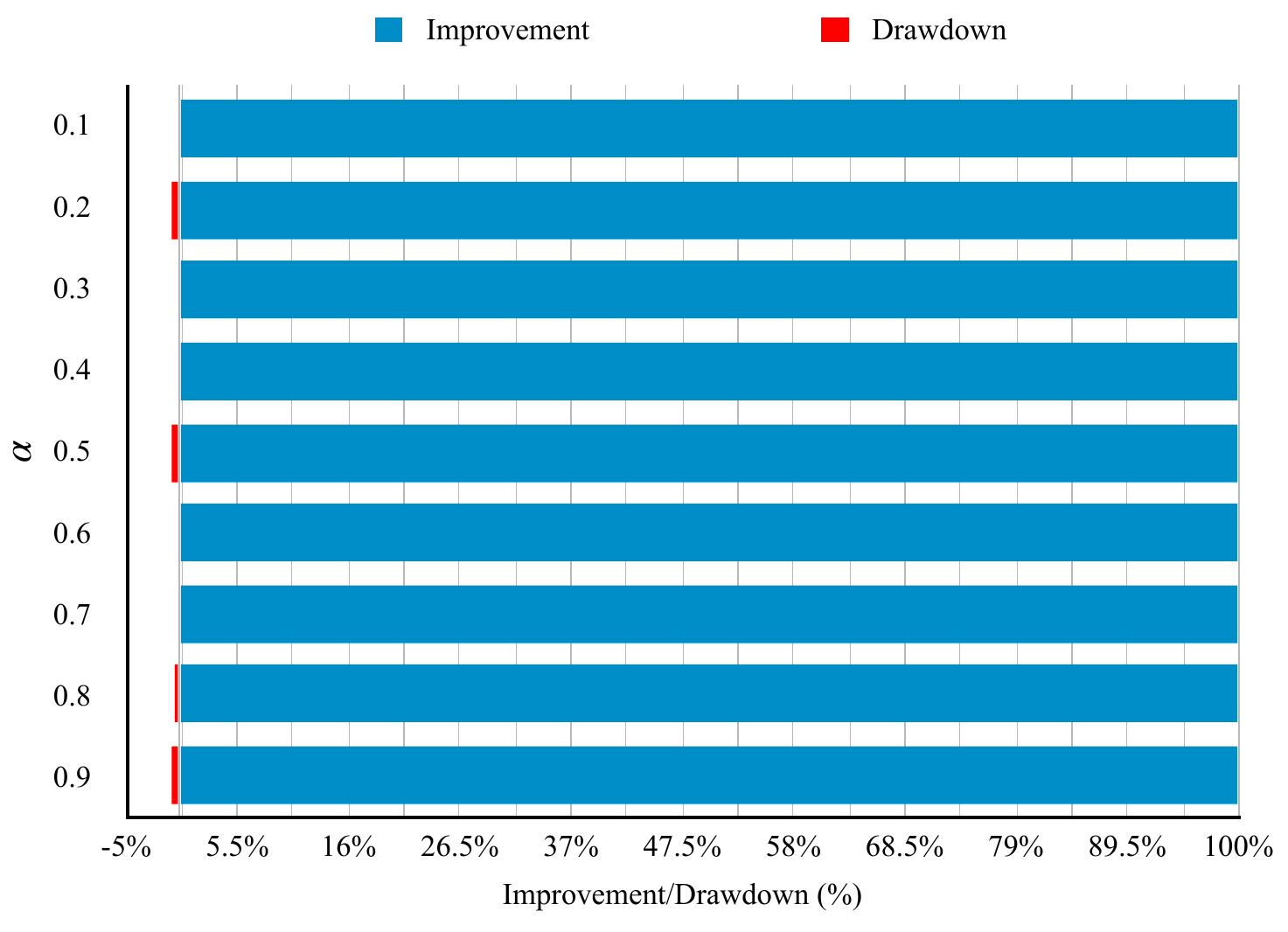}}
    \\
    \subfloat[Comparisons on different repaired neuron numbers.]{\label{repair num}\includegraphics[scale=0.52]{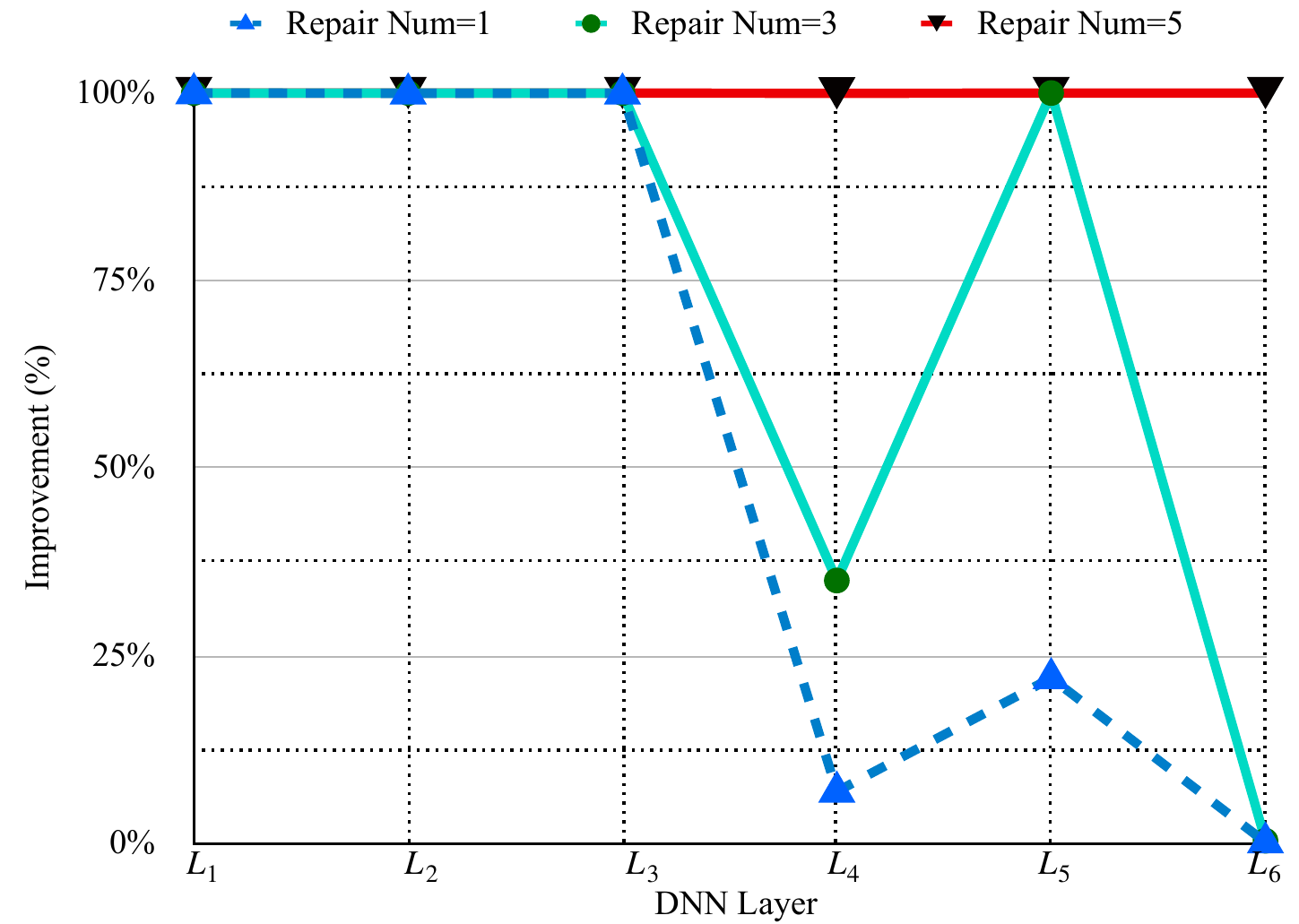}} 
    
    \caption{Performance on different hyperparameters.} 
    \label{hyper compare}
   
\end{figure}

 The results are shown in Fig. \ref{hyper compare}. According to Fig. \ref{alpha}, the weight $\alpha$ balancing the performance drawdown and safety improvement only imposes a slight effect on the final results, which coincides with the conclusions reported in CARE. As for the comparisons among different repaired neuron numbers in Fig. \ref{repair num}, five neurons are enough for every layer and the cases of one and three neurons fail to repair the buggy DNN on some layers, especially on the fourth and the final layers.

\begin{framed}
     \noindent \textit{\textbf{Answer RQ2:} \bird-FT can successfully repair the buggy DNNs, almost without no performance loss, both in cross-layer repair and layer-wise repair styles. Moreover, \bird-FT significantly reduces the running time of state-of-the-art repair works.} 
\end{framed}

\subsection{Activation Compatibility of \bird}
In this subsection,  we make some modifications on the original ACAS Xu DNN \textit{DNN}$_{2}$ to illustrate the compatibility of \bird. That is, we replace the original PWL activation function \texttt{ReLU} with other different activation functions, \texttt{Tanh}, \texttt{LeakyReLU} and \texttt{ELU}, whose definitions and the coefficient values adopted are displayed in Table \ref{tab:actfunc}. Among these \texttt{Tanh} and \texttt{ELU} are non-PWL functions, which means that Veritex and PRDNN cannot tackle them. We term these modified DNNs as \textit{DNN}$_{4}$, \textit{DNN}$_{5}$ and \textit{DNN}$_{6}$, as shown in Table \ref{tab:NN_info}.

\begin{table}[htbp]
    \centering
    \caption{Definitions of Different Activation Functions}
    \label{tab:actfunc}
    
    \setlength{\tabcolsep}{5mm}{
    \begin{tabular}{ccc}
      \Xhline{1.2pt}
       Function &  Definition & Coefficient \\ 
      \hline
      \texttt{ReLU} &  $ \max(x,0)$ & -- \\
      \texttt{Tanh} &  $\frac{e^{x}-e^{-x}}{e^{x}+e^{-x}}$ 
      & -- \\
      \texttt{LeakyReLU} &   \fontsize{6}{6}\selectfont $\begin{cases}
                    x, & x>0\\
                    \alpha x, & x<0
                    \end{cases}$  & $\alpha=0.5$ \\
      \texttt{ELU} &  \fontsize{6}{6}\selectfont  $\begin{cases}
                    x, & x>0\\
                    \alpha(e^{x}-1), & x<0
                    \end{cases}$ & $\alpha=0.5$ \\
     \Xhline{1.2pt}
    \end{tabular}}
\end{table}

\begin{figure}[htbp]
    \centering
    \includegraphics[scale=0.75]{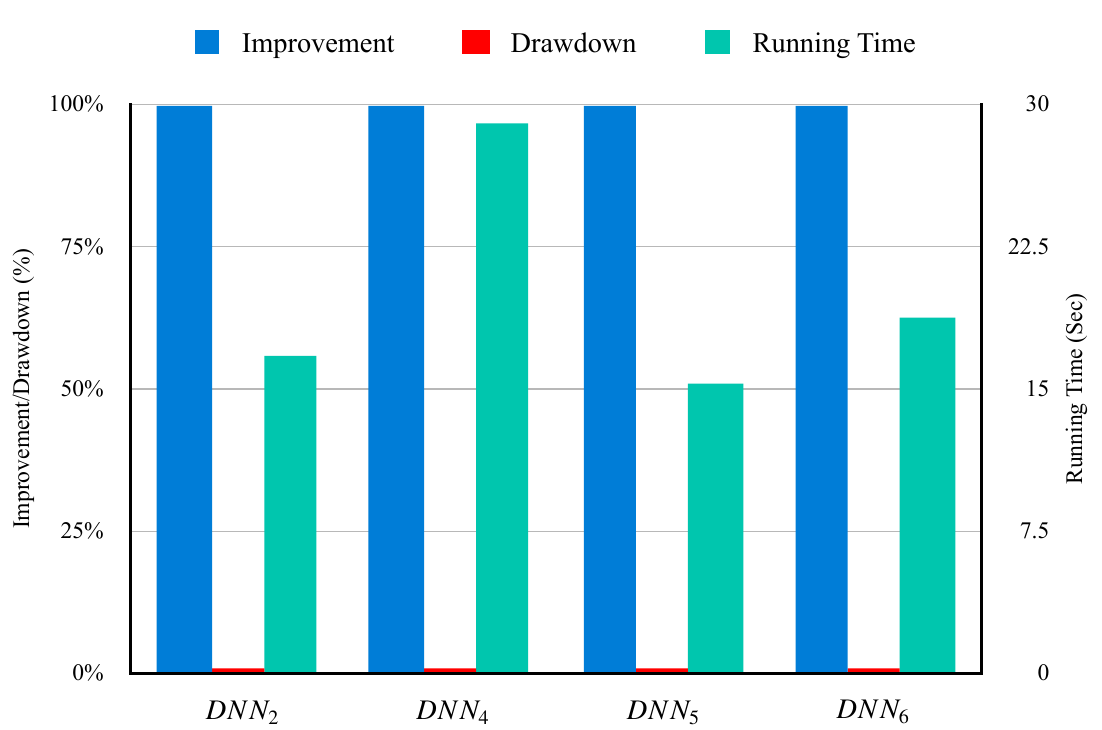}
    \caption{Repair performance of \bird \  on buggy DNNs with different activation functions.}
    \label{fig:different_func}
\end{figure}

Experiment settings follow the ones utilized in Section \ref{evl-ft} and the repair performance of \bird-FT on these DNNs is illustrated in Fig. \ref{fig:different_func}, in terms of safety improvement, performance drawdown and running time. Overall, \bird-FT repairs all the buggy DNNs almost perfectly with little performance drawdown, and the running time is relatively stable. The experimental results demonstrate that \bird \ possesses excellent compatibility with different activation functions, which is not available for Veritex and PRDNN.

\begin{table*}[htbp]
    \centering
    \caption{Comparisons among related work and our proposed repair framework.}
    \label{tab:related_work}
    \setlength{\tabcolsep}{0.8mm}{
    \begin{tabular}{c|cc|ccc|ccc|c}
      \Xhline{1.2pt}
       & ENN\cite{editable2020}   &  Vetitex \cite{yang2022neural} & MMDNN \cite{GoldbergerKAK20}  & LRNN\cite{DBLP:journals/corr/abs-2109-14041} & PRDNN\cite{sotoudeh2021provable}&  NNrepair\cite{DBLP:conf/cav/UsmanGSNP21} & Arachen\cite{DBLP:journals/corr/abs-1912-12463} & CARE\cite{sun2022causality}& \textbf{BIRDNN} (\textbf{Ours}) \\ 
        \Xhline{1.2pt}
\rowcolor{gray!30}     Retraining  & \checkmark & \checkmark & & & & & & & \checkmark\\
     Fine-tuning  & & &\checkmark &\checkmark & \checkmark& \checkmark &\checkmark & \checkmark& \checkmark\\
  \rowcolor{gray!30}   Fault Localization  & & & & & &\checkmark & \checkmark& \checkmark &\checkmark\\
          
     Provable Repair$^{*}$  & 
    & $\triangle$ & & & \checkmark & & & & $\triangle$ \\
  \rowcolor{gray!30}  Activation Compatibility & \checkmark &  &  &&&  & \checkmark & \checkmark & \checkmark \\
     DRPs$^{\dag}$  &  & \checkmark & & \checkmark & \checkmark& & & \checkmark& \checkmark\\
       \Xhline{1.2pt}
    \end{tabular}}
    \begin{tablenotes}    
        \footnotesize               
       \item \textbf{*}  Veritex and BIRDNN (retraining)  are provable repair on condition that the loss function value reaches 0 and we mark them with $\triangle$, instead of \checkmark.
       \item \textbf{$\dag$} The repair scenarios tackled by each work are identified from the shown evaluation tests, and some may work for the untested cases with modifications.
      \end{tablenotes}
\end{table*}

\begin{framed}
     \noindent \textit{\textbf{Answer RQ3:} Compared with the works like Veritex and PRDNN, \bird \ owns great compatibility with different activation functions with little performance loss.} 
\end{framed}

\subsection{Threats to Validity}
The primary threat is that \bird \  is essentially a probably approximated correct repair framework. It provides a probabilistic guarantee for DRPs, and this guarantee's confidence is directly related to the sample size.

Another possible threat is the overfitting risk resulting from sample based repair, which widely exists in previous methods. Therefore,  different and large-size training/repairing data and test data are adopted to avoid overfitting and evaluate the repaired DNNs' generalization, and we follow their setups.


\section{Related Work}
\label{related}

DNN repair has received significant attention in recent decades, and dozens of works on DNN repair have witnessed remarkable advances. In what follows, we introduce the related work from three categories: retraining, fine-tuning without fault localization, and fine-tuning with fault localization.

\textbf{Retraining.} 
Veritex \cite{yang2022neural} is one of the most relevant works to our research and it presents an approach to repairing unsafe ReLU DNNs via reachability analysis. To patch DNNs against the violated safety properties, the retraining process of Veritex utilizes a loss function to constrain the distance between the exact unsafe output region and the safe domain. Additionally, an extra loss term is combined to minimize the DNN patch and avoid introducing new unexpected behaviors synergistically. In contrast, we assign negative samples with expected labels via behavior imitation, which is more lightweight. Before that, \cite{editable2020} addresses DNN repair problems by introducing a gradient-based and model-agnostic editable training strategy, named editable neural networks (ENNs).

\textbf{Fine-tuning without fault localization.} Resorting to DNN verification techniques, \cite{GoldbergerKAK20} utilizes the recent advances to patch a faulty DNN to satisfy specific requirements in a provable minimal manner. Additionally, \cite{DBLP:journals/corr/abs-2109-14041} presents a method to repair feed-forward deep neural networks via reducing the predicate satisfaction formulated from the patch problem into a mixed integer quadratic program (MIQP) to calibrate the weights of a single layer. Furthermore, PRDNN \cite{sotoudeh2021provable}, another related work to ours, introduces a novel and equivalent representation of the buggy ReLU DNNs, Decouple DNNs (DDNNs), reducing the DNN repair problem to a linear programming (LP) problem. However, the repair process works aimlessly and modifies the parameters of an arbitrary DNN layer, without fault localization preprocessing. Also, due to the expensive complexity of computing linear polytope regions of DNNs, the dimension of the repaired polytopes is limited, and generally, only one- or two-dimension is feasible. Besides, converting the resultant DDNNs back into standard and equivalent DNNs is largely an open problem.

\textbf{Fine-tuning with fault localization.}
Closely resembling finding out the erroneous codes in software engineering, fault localization identifies the suspicious (responsible)  neuron set responsible for the unexpected DNN behaviors. DeepFault \cite{DBLP:conf/fase/EniserGS19} is the first proposed work on fault localization of DNNs, which analyzes pre-trained DNNs against a test set to establish the hit spectrum corresponding to each neuron, and then the suspicious neurons are identified by employing a designed suspiciousness measure. Whereas, instead of DNN repair, this fault localization technique is then utilized to synthesize adversarial inputs. Subsequently, NNrepair \cite{DBLP:conf/cav/UsmanGSNP21} explores the activation patterns \cite{DBLP:conf/kbse/GopinathCPT19} for the identification of potentially faulty neurons and the repair constraints w.r.t. DNN parameters are solved via concolic execution \cite{DBLP:conf/sigsoft/SenMA05}. Further,
Arachen \cite{DBLP:journals/corr/abs-1912-12463} localizes the neural weights which have more impact on negative samples and less impact on positive samples via bidirectional localization, inspired by the fault localization of software engineering. In the following, the selected neuron weights are modified with the differential evolution algorithm. Then it comes to another related work to this paper, CARE \cite{sun2022causality}, which follows the same patch paradigm as Arachen, while, it performs a causality-based fault localization for identifying the responsible neurons and its parameter optimization strategy adopts the particle swarm optimization (PSO) algorithm. Nevertheless, the fault localization strategy is 
less efficient due to the computation of the causality operators.

In summary, the comprehensive comparisons among these related work and  \bird \ are demonstrated in Table \ref{tab:related_work}. 

\section{Conclusion}
\label{concl}

This paper makes a unique insight on the behavior differences of DNN neurons and proposes a DNN repair framework \bird \ based on behavior imitation, which supports alternative retraining based and fine-tuning based  DNN patching for the first time. Besides, \bird \  tackles domain-wise repair problems in a probably approximated correct style by characterizing DNN behaviors over domains. Experiments on the ACAS Xu DNNs illustrate \bird's effectiveness, efficiency and compatibility. Despite the outstanding performance, improving \bird \  further to be a provable DNN repair framework is an appealing and challenging future work. A deeper investigation into the behavior differences of DNNs may provide specific directions to modify the neurons (adding or subtracting some values). Additionally, utilizing the proposed fault localization strategy to synthesize adversarial samples is another possible future work.

 \bibliographystyle{IEEEbib} 
 \bibliography{reference}

\begin{thebibliography}{10}
\providecommand{\url}[1]{#1}
\csname url@samestyle\endcsname
\providecommand{\newblock}{\relax}
\providecommand{\bibinfo}[2]{#2}
\providecommand{\BIBentrySTDinterwordspacing}{\spaceskip=0pt\relax}
\providecommand{\BIBentryALTinterwordstretchfactor}{4}
\providecommand{\BIBentryALTinterwordspacing}{\spaceskip=\fontdimen2\font plus
\BIBentryALTinterwordstretchfactor\fontdimen3\font minus
  \fontdimen4\font\relax}
\providecommand{\BIBforeignlanguage}[2]{{%
\expandafter\ifx\csname l@#1\endcsname\relax
\typeout{** WARNING: IEEEtran.bst: No hyphenation pattern has been}%
\typeout{** loaded for the language `#1'. Using the pattern for}%
\typeout{** the default language instead.}%
\else
\language=\csname l@#1\endcsname
\fi
#2}}
\providecommand{\BIBdecl}{\relax}
\BIBdecl

\bibitem{dahnert2021panoptic}
M.~Dahnert, J.~Hou, M.~Nie{\ss}ner, and A.~Dai, ``Panoptic 3d scene
  reconstruction from a single rgb image,'' \emph{Advances in Neural
  Information Processing Systems}, vol.~34, 2021.

\bibitem{yuan2021bartscore}
W.~Yuan, G.~Neubig, and P.~Liu, ``Bartscore: Evaluating generated text as text
  generation,'' \emph{arXiv preprint arXiv:2106.11520}, 2021.

\bibitem{karch2021grounding}
T.~Karch, L.~Teodorescu, K.~Hofmann, C.~Moulin-Frier, and P.-Y. Oudeyer,
  ``Grounding spatio-temporal language with transformers,'' \emph{arXiv
  preprint arXiv:2106.08858}, 2021.

\bibitem{DBLP:journals/asc/XuXTH03}
\BIBentryALTinterwordspacing
K.~Xu, M.~Xie, L.~C. Tang, and S.~L. Ho, ``Application of neural networks in
  forecasting engine systems reliability,'' \emph{Appl. Soft Comput.}, vol.~2,
  no.~4, pp. 255--268, 2003. [Online]. Available:
  \url{https://doi.org/10.1016/S1568-4946(02)00059-5}
\BIBentrySTDinterwordspacing

\bibitem{DBLP:conf/IEEEicci/Berwick21}
R.~C. Berwick, ``The failure of deep neural networks to capture human
  language's cognitive core,'' in \emph{20th {IEEE} International Conference on
  Cognitive Informatics {\&} Cognitive Computing, ICCI*CC 2021}.\hskip 1em plus
  0.5em minus 0.4em\relax {IEEE}, 2021, p.~3.

\bibitem{DBLP:conf/latw/BosioBR019}
\BIBentryALTinterwordspacing
A.~Bosio, P.~Bernardi, A.~Ruospo, and E.~S{\'{a}}nchez, ``A reliability
  analysis of a deep neural network,'' in \emph{{IEEE} Latin American Test
  Symposium, {LATS} 2019, Santiago, Chile, March 11-13, 2019}.\hskip 1em plus
  0.5em minus 0.4em\relax {IEEE}, 2019, pp. 1--6. [Online]. Available:
  \url{https://doi.org/10.1109/LATW.2019.8704548}
\BIBentrySTDinterwordspacing

\bibitem{translationarrest}
A.~Hern, ``Facebook translates 'good morning' into 'attack them', leading to
  arrest,''
  \url{https://www.theguardian.com/technology/2017/oct/24/facebook-palestine-israel-translates-good-morning-attack-them-arrest},
  2017, accessed: 2020-06-06.

\bibitem{wronglyaccused}
K.~Hill, ``Wrongfully accused by an algorithm,'' {New York Times.}
  \url{https://www.nytimes.com/2020/06/24/technology/facial-recognition-arrest.html},
  2020, accessed: 2020-06-06.

\bibitem{teslacrash}
D.~Lee, ``{US} opens investigation into {Tesla} after fatal crash,'' {BBC.}
  \url{https://www.bbc.co.uk/news/technology-36680043}, 2016, accessed:
  2020-06-06.

\bibitem{2021Enhancing}
P.~Yang, J.~Li, J.~Liu, C.~C. Huang, R.~Li, L.~Chen, X.~Huang, and L.~Zhang,
  ``Enhancing robustness verification for deep neural networks viasymbolic
  propagation,'' \emph{Formal Aspects of Computing}, vol.~33, no.~3, pp.
  407--435, 2021.

\bibitem{DBLP:journals/jcst/LiuSZW20}
W.~Liu, F.~Song, T.~Zhang, and J.~Wang, ``Verifying relu neural networks from a
  model checking perspective,'' \emph{J. Comput. Sci. Technol.}, vol.~35,
  no.~6, pp. 1365--1381, 2020.

\bibitem{goodfellow2014explaining}
I.~J. Goodfellow, J.~Shlens, and C.~Szegedy, ``Explaining and harnessing
  adversarial examples,'' \emph{arXiv preprint arXiv:1412.6572}, 2014.

\bibitem{liang2022safety}
Z.~Liang, D.~Ren, W.~Liu, J.~Wang, W.~Yang, and B.~Xue, ``Safety verification
  for neural networks based on set-boundary analysis,'' \emph{arXiv preprint
  arXiv:2210.04175}, 2022.

\bibitem{DBLP:journals/pacmpl/SinghGPV19}
G.~Singh, T.~Gehr, M.~P{\"{u}}schel, and M.~T. Vechev, ``An abstract domain for
  certifying neural networks,'' \emph{Proc. {ACM} Program. Lang.}, vol.~3, no.
  {POPL}, pp. 41:1--41:30, 2019.

\bibitem{DBLP:conf/tacas/YangLLHWSXZ21}
P.~Yang, R.~Li, J.~Li, C.~Huang, J.~Wang, J.~Sun, B.~Xue, and L.~Zhang,
  ``Improving neural network verification through spurious region guided
  refinement,'' in \emph{Tools and Algorithms for the Construction and Analysis
  of Systems - 27th International Conference, {TACAS} 2021}, ser. Lecture Notes
  in Computer Science, vol. 12651.\hskip 1em plus 0.5em minus 0.4em\relax
  Springer, 2021, pp. 389--408.

\bibitem{DBLP:conf/sp/GehrMDTCV18}
T.~Gehr, M.~Mirman, D.~Drachsler{-}Cohen, P.~Tsankov, S.~Chaudhuri, and M.~T.
  Vechev, ``{AI2:} safety and robustness certification of neural networks with
  abstract interpretation,'' in \emph{2018 {IEEE} Symposium on Security and
  Privacy, {SP} 2018}.\hskip 1em plus 0.5em minus 0.4em\relax {IEEE} Computer
  Society, 2018, pp. 3--18.

\bibitem{DBLP:conf/icml/KoLWDWL19}
C.~Ko, Z.~Lyu, L.~Weng, L.~Daniel, N.~Wong, and D.~Lin, ``{POPQORN:}
  quantifying robustness of recurrent neural networks,'' in \emph{Proceedings
  of the 36th International Conference on Machine Learning, {ICML} 2019}, ser.
  Proceedings of Machine Learning Research, vol.~97.\hskip 1em plus 0.5em minus
  0.4em\relax {PMLR}, 2019, pp. 3468--3477.

\bibitem{DBLP:conf/cav/KatzBDJK17}
G.~Katz, C.~W. Barrett, D.~L. Dill, K.~Julian, and M.~J. Kochenderfer,
  ``Reluplex: An efficient {SMT} solver for verifying deep neural networks,''
  in \emph{Computer Aided Verification - 29th International Conference, {CAV}
  2017}, ser. Lecture Notes in Computer Science, vol. 10426.\hskip 1em plus
  0.5em minus 0.4em\relax Springer, 2017, pp. 97--117.

\bibitem{DBLP:conf/tacas/Amir0BK21}
G.~Amir, H.~Wu, C.~W. Barrett, and G.~Katz, ``An smt-based approach for
  verifying binarized neural networks,'' in \emph{Tools and Algorithms for the
  Construction and Analysis of Systems - 27th International Conference, {TACAS}
  2021}, ser. Lecture Notes in Computer Science, vol. 12652.\hskip 1em plus
  0.5em minus 0.4em\relax Springer, 2021, pp. 203--222.

\bibitem{DBLP:conf/iclr/TjengXT19}
V.~Tjeng, K.~Y. Xiao, and R.~Tedrake, ``Evaluating robustness of neural
  networks with mixed integer programming,'' in \emph{7th International
  Conference on Learning Representations, {ICLR} 2019}.\hskip 1em plus 0.5em
  minus 0.4em\relax OpenReview.net, 2019.

\bibitem{DBLP:conf/ijcai/BattenKLZ21}
B.~Batten, P.~Kouvaros, A.~Lomuscio, and Y.~Zheng, ``Efficient neural network
  verification via layer-based semidefinite relaxations and linear cuts,'' in
  \emph{Proceedings of the Thirtieth International Joint Conference on
  Artificial Intelligence, {IJCAI} 2021}.\hskip 1em plus 0.5em minus
  0.4em\relax ijcai.org, 2021, pp. 2184--2190.

\bibitem{DBLP:conf/cvpr/LinYCZLLH19}
W.~Lin, Z.~Yang, X.~Chen, Q.~Zhao, X.~Li, Z.~Liu, and J.~He, ``Robustness
  verification of classification deep neural networks via linear programming,''
  in \emph{{IEEE} Conference on Computer Vision and Pattern Recognition, {CVPR}
  2019}.\hskip 1em plus 0.5em minus 0.4em\relax Computer Vision Foundation /
  {IEEE}, 2019, pp. 11\,418--11\,427.

\bibitem{DBLP:conf/qrs/ShenWC18}
W.~Shen, J.~Wan, and Z.~Chen, ``Munn: Mutation analysis of neural networks,''
  in \emph{2018 {IEEE} International Conference on Software Quality,
  Reliability and Security Companion, {QRS} Companion 2018}.\hskip 1em plus
  0.5em minus 0.4em\relax {IEEE}, 2018, pp. 108--115.

\bibitem{DBLP:conf/icse/TianPJR18}
Y.~Tian, K.~Pei, S.~Jana, and B.~Ray, ``Deeptest: automated testing of
  deep-neural-network-driven autonomous cars,'' in \emph{Proceedings of the
  40th International Conference on Software Engineering, {ICSE} 2018}.\hskip
  1em plus 0.5em minus 0.4em\relax {ACM}, 2018, pp. 303--314.

\bibitem{DBLP:conf/issta/XieMJXCLZLYS19}
X.~Xie, L.~Ma, F.~Juefei{-}Xu, M.~Xue, H.~Chen, Y.~Liu, J.~Zhao, B.~Li, J.~Yin,
  and S.~See, ``Deephunter: a coverage-guided fuzz testing framework for deep
  neural networks,'' in \emph{Proceedings of the 28th {ACM} {SIGSOFT}
  International Symposium on Software Testing and Analysis, {ISSTA}
  2019}.\hskip 1em plus 0.5em minus 0.4em\relax {ACM}, 2019, pp. 146--157.

\bibitem{DBLP:conf/icse/ZhangXMDH0Z020}
X.~Zhang, X.~Xie, L.~Ma, X.~Du, Q.~Hu, Y.~Liu, J.~Zhao, and M.~Sun, ``Towards
  characterizing adversarial defects of deep learning software from the lens of
  uncertainty,'' in \emph{{ICSE} '20: 42nd International Conference on Software
  Engineering}.\hskip 1em plus 0.5em minus 0.4em\relax {ACM}, 2020, pp.
  739--751.

\bibitem{DBLP:journals/corr/abs-2005-00760}
Z.~Chen, Y.~Cao, Y.~Liu, H.~Wang, T.~Xie, and X.~Liu, ``Understanding
  challenges in deploying deep learning based software: An empirical study,''
  \emph{CoRR}, vol. abs/2005.00760, 2020.

\bibitem{gowal2018effectiveness}
S.~Gowal, K.~Dvijotham, R.~Stanforth, R.~Bunel, C.~Qin, J.~Uesato, T.~Mann, and
  P.~Kohli, ``On the effectiveness of interval bound propagation for training
  verifiably robust models,'' \emph{CoRR}, vol. abs/1810.12715, 2018.

\bibitem{madry2017towards}
A.~Madry, A.~Makelov, L.~Schmidt, D.~Tsipras, and A.~Vladu, ``Towards deep
  learning models resistant to adversarial attacks,'' \emph{CoRR}, vol.
  abs/1706.06083, 2017.

\bibitem{yang2021reachability}
X.~Yang, T.~T. Johnson, H.-D. Tran, T.~Yamaguchi, B.~Hoxha, and D.~V.
  Prokhorov, ``Reachability analysis of deep relu neural networks using
  facet-vertex incidence,'' in \emph{HSCC}, vol.~21, 2021, pp. 19--21.

\bibitem{sinitsin2020editable}
A.~Sinitsin, V.~Plokhotnyuk, D.~Pyrkin, S.~Popov, and A.~Babenko, ``Editable
  neural networks,'' \emph{arXiv preprint arXiv:2004.00345}, 2020.

\bibitem{tao2023architecture}
Z.~Tao, S.~Nawas, J.~Mitchell, and A.~V. Thakur, ``Architecture-preserving
  provable repair of deep neural networks,'' \emph{arXiv preprint
  arXiv:2304.03496}, 2023.

\bibitem{GoldbergerKAK20}
B.~Goldberger, G.~Katz, Y.~Adi, and J.~Keshet, ``Minimal modifications of deep
  neural networks using verification,'' in \emph{{LPAR} 2020: 23rd
  International Conference on Logic for Programming, Artificial Intelligence
  and Reasoning}, ser. EPiC Series in Computing, vol.~73, 2020, pp. 260--278.

\bibitem{sotoudeh2021provable}
M.~Sotoudeh and A.~V. Thakur, ``Provable repair of deep neural networks,'' in
  \emph{Proceedings of the 42nd ACM SIGPLAN International Conference on
  Programming Language Design and Implementation}, 2021, pp. 588--603.

\bibitem{DBLP:conf/cav/UsmanGSNP21}
M.~Usman, D.~Gopinath, Y.~Sun, Y.~Noller, and C.~S. Pasareanu, ``Nnrepair:
  Constraint-based repair of neural network classifiers,'' in \emph{Computer
  Aided Verification - 33rd International Conference, {CAV} 2021}, ser. Lecture
  Notes in Computer Science, vol. 12759.\hskip 1em plus 0.5em minus 0.4em\relax
  Springer, 2021, pp. 3--25.

\bibitem{sun2022causality}
B.~Sun, J.~Sun, L.~H. Pham, and J.~Shi, ``Causality-based neural network
  repair,'' in \emph{Proceedings of the 44th International Conference on
  Software Engineering}, 2022, pp. 338--349.

\bibitem{mitchell2007machine}
T.~M. Mitchell \emph{et~al.}, \emph{Machine learning}.\hskip 1em plus 0.5em
  minus 0.4em\relax McGraw-hill New York, 2007, vol.~1.

\bibitem{casadio2022neural}
M.~Casadio, E.~Komendantskaya, M.~L. Daggitt, W.~Kokke, G.~Katz, G.~Amir, and
  I.~Refaeli, ``Neural network robustness as a verification property: a
  principled case study,'' in \emph{Computer Aided Verification: 34th
  International Conference, CAV 2022, Haifa, Israel, August 7--10, 2022,
  Proceedings, Part I}.\hskip 1em plus 0.5em minus 0.4em\relax Springer, 2022,
  pp. 219--231.

\bibitem{huang2019reachnn}
C.~Huang, J.~Fan, W.~Li, X.~Chen, and Q.~Zhu, ``Reachnn: Reachability analysis
  of neural-network controlled systems,'' \emph{ACM Transactions on Embedded
  Computing Systems (TECS)}, vol.~18, no.~5s, pp. 1--22, 2019.

\bibitem{sun2021probabilistic}
B.~Sun, J.~Sun, T.~Dai, and L.~Zhang, ``Probabilistic verification of neural
  networks against group fairness,'' in \emph{Formal Methods: 24th
  International Symposium, FM 2021, Virtual Event, November 20--26, 2021,
  Proceedings 24}.\hskip 1em plus 0.5em minus 0.4em\relax Springer, 2021, pp.
  83--102.

\bibitem{fischer2019dl2}
M.~Fischer, M.~Balunovic, D.~Drachsler-Cohen, T.~Gehr, C.~Zhang, and M.~Vechev,
  ``Dl2: training and querying neural networks with logic,'' in
  \emph{International Conference on Machine Learning}.\hskip 1em plus 0.5em
  minus 0.4em\relax PMLR, 2019, pp. 1931--1941.

\bibitem{liu2020verifying}
W.-W. Liu, F.~Song, T.-H.-R. Zhang, and J.~Wang, ``Verifying relu neural
  networks from a model checking perspective,'' \emph{Journal of Computer
  Science and Technology}, vol.~35, pp. 1365--1381, 2020.

\bibitem{sohn2022arachne}
J.~Sohn, S.~Kang, and S.~Yoo, ``Arachne: Search based repair of deep neural
  networks,'' \emph{ACM Transactions on Software Engineering and Methodology},
  2022.

\bibitem{DBLP:series/lncs/LegayLTYSG19}
A.~Legay, A.~Lukina, L.~Traonouez, J.~Yang, S.~A. Smolka, and R.~Grosu,
  ``Statistical model checking,'' in \emph{Computing and Software Science -
  State of the Art and Perspectives}, ser. Lecture Notes in Computer
  Science.\hskip 1em plus 0.5em minus 0.4em\relax Springer, 2019, vol. 10000,
  pp. 478--504.

\bibitem{agha2018survey}
G.~Agha and K.~Palmskog, ``A survey of statistical model checking,'' \emph{ACM
  Transactions on Modeling and Computer Simulation (TOMACS)}, vol.~28, no.~1,
  pp. 1--39, 2018.

\bibitem{1959Some}
M.~Okamoto, ``Some inequalities relating to the partial sum of binomial
  probabilities,'' \emph{Annals of the Institute of Statistical Mathematics},
  vol.~10, no.~1, pp. 29--35, 1959.

\bibitem{DBLP:conf/icse/LiYHS0Z22}
R.~Li, P.~Yang, C.~Huang, Y.~Sun, B.~Xue, and L.~Zhang, ``Towards practical
  robustness analysis for dnns based on pac-model learning,'' in \emph{44th
  {IEEE/ACM} 44th International Conference on Software Engineering, {ICSE}
  2022, Pittsburgh, PA, USA, May 25-27, 2022}.\hskip 1em plus 0.5em minus
  0.4em\relax {ACM}, 2022, pp. 2189--2201.

\bibitem{kennedy1995particle}
J.~Kennedy and R.~Eberhart, ``Particle swarm optimization,'' in
  \emph{Proceedings of ICNN'95-international conference on neural networks},
  vol.~4.\hskip 1em plus 0.5em minus 0.4em\relax IEEE, 1995, pp. 1942--1948.

\bibitem{storn1997differential}
R.~Storn and K.~Price, ``Differential evolution-a simple and efficient
  heuristic for global optimization over continuous spaces,'' \emph{Journal of
  global optimization}, vol.~11, no.~4, p. 341, 1997.

\bibitem{fok2017optimization}
R.~Fok, A.~An, and X.~Wang, ``Optimization assisted mcmc,'' \emph{arXiv
  preprint arXiv:1709.02888}, 2017.

\bibitem{trojovsky2022pelican}
P.~Trojovsk{\`y} and M.~Dehghani, ``Pelican optimization algorithm: A novel
  nature-inspired algorithm for engineering applications,'' \emph{Sensors},
  vol.~22, no.~3, p. 855, 2022.

\bibitem{shi1998parameter}
Y.~Shi and R.~C. Eberhart, ``Parameter selection in particle swarm
  optimization,'' in \emph{Evolutionary Programming VII: 7th International
  Conference}.\hskip 1em plus 0.5em minus 0.4em\relax Springer, 1998, pp.
  591--600.

\bibitem{DBLP:journals/corr/abs-1810-04240}
K.~D. Julian, M.~J. Kochenderfer, and M.~P. Owen, ``Deep neural network
  compression for aircraft collision avoidance systems,'' \emph{CoRR}, vol.
  abs/1810.04240, 2018.

\bibitem{DBLP:journals/corr/abs-2007-11206}
L.~H. Pham, J.~Li, and J.~Sun, ``{SOCRATES:} towards a unified platform for
  neural network verification,'' \emph{CoRR}, vol. abs/2007.11206, 2020.

\bibitem{yang2022neural}
X.~Yang, T.~Yamaguchi, H.-D. Tran, B.~Hoxha, T.~T. Johnson, and D.~Prokhorov,
  ``Neural network repair with reachability analysis,'' in \emph{Formal
  Modeling and Analysis of Timed Systems: 20th International Conference,
  FORMATS 2022}.\hskip 1em plus 0.5em minus 0.4em\relax Springer, 2022, pp.
  221--236.

\bibitem{DBLP:conf/fmcad/LinZSJ20}
X.~Lin, H.~Zhu, R.~Samanta, and S.~Jagannathan, ``Art: Abstraction
  refinement-guided training for provably correct neural networks,'' in
  \emph{2020 Formal Methods in Computer Aided Design, {FMCAD} 2020, Haifa,
  Israel, September 21-24, 2020}.\hskip 1em plus 0.5em minus 0.4em\relax
  {IEEE}, 2020, pp. 148--157.

\bibitem{DBLP:journals/swarm/PoliKB07}
R.~Poli, J.~Kennedy, and T.~Blackwell, ``Particle swarm optimization,''
  \emph{Swarm Intell.}, vol.~1, no.~1, pp. 33--57, 2007.

\bibitem{editable2020}
A.~Sinitsin, V.~Plokhotnyuk, D.~V. Pyrkin, S.~Popov, and A.~Babenko, ``Editable
  neural networks,'' in \emph{8th International Conference on Learning
  Representations, {ICLR} 2020}, 2020.

\bibitem{DBLP:journals/corr/abs-2109-14041}
K.~Majd, S.~Zhou, H.~B. Amor, G.~Fainekos, and S.~Sankaranarayanan, ``Local
  repair of neural networks using optimization,'' \emph{CoRR}, vol.
  abs/2109.14041, 2021.

\bibitem{DBLP:journals/corr/abs-1912-12463}
\BIBentryALTinterwordspacing
J.~Sohn, S.~Kang, and S.~Yoo, ``Search based repair of deep neural networks,''
  \emph{CoRR}, vol. abs/1912.12463, 2019. [Online]. Available:
  \url{http://arxiv.org/abs/1912.12463}
\BIBentrySTDinterwordspacing

\bibitem{DBLP:conf/fase/EniserGS19}
H.~F. Eniser, S.~Gerasimou, and A.~Sen, ``Deepfault: Fault localization for
  deep neural networks,'' in \emph{Fundamental Approaches to Software
  Engineering - 22nd International Conference, {FASE} 2019}, ser. Lecture Notes
  in Computer Science, vol. 11424.\hskip 1em plus 0.5em minus 0.4em\relax
  Springer, 2019, pp. 171--191.

\bibitem{DBLP:conf/kbse/GopinathCPT19}
D.~Gopinath, H.~Converse, C.~S. Pasareanu, and A.~Taly, ``Property inference
  for deep neural networks,'' in \emph{34th {IEEE/ACM} International Conference
  on Automated Software Engineering, {ASE} 2019}.\hskip 1em plus 0.5em minus
  0.4em\relax {IEEE}, 2019, pp. 797--809.

\bibitem{DBLP:conf/sigsoft/SenMA05}
K.~Sen, D.~Marinov, and G.~Agha, ``{CUTE:} a concolic unit testing engine for
  {C},'' in \emph{Proceedings of the 10th European Software Engineering
  Conference held jointly with 13th {ACM} {SIGSOFT} International Symposium on
  Foundations of Software Engineering, 2005}.\hskip 1em plus 0.5em minus
  0.4em\relax {ACM}, 2005, pp. 263--272.

\end{thebibliography}
\end{document}